\documentclass{article}

    \PassOptionsToPackage{numbers, compress}{natbib}
\usepackage[final]{neurips_2024}

\usepackage[utf8]{inputenc} % allow utf-8 input
\usepackage[T1]{fontenc}    % use 8-bit T1 fonts
\usepackage{hyperref}       % hyperlinks
\usepackage{url}            % simple URL typesetting
\usepackage{booktabs}       % professional-quality tables
\usepackage{amsfonts}       % blackboard math symbols
\usepackage{nicefrac}       % compact symbols for 1/2, etc.
\usepackage{microtype}      % microtypography
\usepackage{xcolor}         % colors

\usepackage{graphicx}
\usepackage{subfigure}
\usepackage{amsthm}

\theoremstyle{definition}
\newtheorem{theorem}{Theorem}

\newtheorem{assumption}{Assumption}

\usepackage{booktabs}
\usepackage{caption}
\usepackage{wrapfig}
\usepackage{amsmath}
\usepackage{amssymb}
\usepackage{pifont}
\newcommand{\cmark}{\ding{51}}%
\newcommand{\xmark}{\ding{55}}%
\usepackage{float}

\usepackage{enumitem}
\setlist[itemize]{leftmargin=*} % To remove the indentation

\usepackage{algorithm2e}
\RestyleAlgo{ruled}
\SetKwInput{KwData}{Input}
\SetKwInput{KwResult}{Output}

\usepackage[capitalize]{cleveref}
\crefname{assumption}{assumption}{assumptions}

\title{Mitigating Spurious Correlations \\ via Disagreement Probability}
\author{%
Hyeonggeun Han$^{1,2}$\quad Sehwan Kim$^{1}$\quad Hyungjun Joo$^{1,2}$ \\ \textbf{Sangwoo Hong}$^{1,2}$\quad \textbf{Jungwoo Lee}$^{1,2,3}$\thanks{Corresponding author} \\
$^1$ECE \& $^2$NextQuantum, Seoul National University \quad $^3$Hodoo AI Labs \\
\texttt{\{hygnhan, sehwankim, joohj911, tkddn0606, junglee\}@snu.ac.kr}\\
}

\begin{document}

\maketitle

\begin{abstract}
Models trained with empirical risk minimization (ERM) are prone to be biased towards spurious correlations between target labels and bias attributes, which leads to poor performance on data groups lacking spurious correlations. It is particularly challenging to address this problem when access to bias labels is not permitted. To mitigate the effect of spurious correlations without bias labels, we first introduce a novel training objective designed to robustly enhance model performance across all data samples, irrespective of the presence of spurious correlations. From this objective, we then derive a debiasing method, Disagreement Probability based Resampling for debiasing (DPR), which does not require bias labels. DPR leverages the disagreement between the target label and the prediction of a biased model to identify bias-conflicting samples—those without spurious correlations—and upsamples them according to the disagreement probability. Empirical evaluations on multiple benchmarks demonstrate that DPR achieves state-of-the-art performance over existing baselines that do not use bias labels. Furthermore, we provide a theoretical analysis that details how DPR reduces dependency on spurious correlations.
\end{abstract}

\section{Introduction}
\begin{wrapfigure}{h}{0.32\columnwidth}
    \vspace{-6.0mm}
\centerline{\includegraphics[width=0.32\columnwidth]{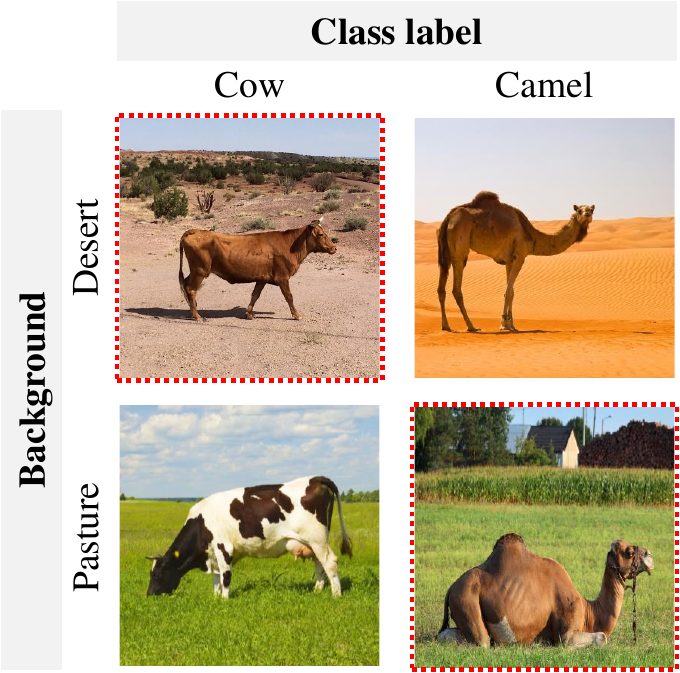}}
    \caption{An illustration of the cow/camel classification task. Red dotted boxes indicate samples where spurious correlations do not hold.}
    \label{fig:cow_camel}
    \vspace{-4.0mm}
\end{wrapfigure}
% para1: spurious correlation 이 무엇인지, 어떤 영향을 주는지 (위험성), 왜 중요한지 (왜 해결해야하는지, 이는 보통 잘 일어나는 현상이다 등등 첨언하면서) (구체적인 예시를 들면 더 좋을듯. 이 때 이해를 돕기 위한 시각자료 (figure) 하나 넣으면 더 좋지 않을까? spurious correlation 설명하기 좋은) (NN은 보통 ERM으로 학습시킨다. 그러나 ERM은 spurious correlation에 취약하다. 등등 고려하여 작성하기.)
In the realm of machine learning, many classification models employ Empirical Risk Minimization (ERM) \cite{ERM}, which optimizes average performance. However, this approach has been found to underperform on certain groups of data \cite{blodgett2016,hashimoto2018,duchi2019} due to the prevalence of spurious correlations within training datasets \cite{mccoy2019,groupDRO,sagawa2020}. Spurious correlations refer to the strong correlations between target labels and easy-to-learn attributes (\textit{i.e.}, bias attributes), which are present in a majority of the training examples. ERM-trained models often rely on these bias attributes \cite{geirhos2020}, leading to biased predictions and poor generalization on minority groups where spurious correlations are absent. For example, consider the cow/camel classification task illustrated in \Cref{fig:cow_camel}. A majority of camel images feature desert backgrounds, while a majority of cow images feature pasture backgrounds. Models trained via ERM might learn to recognize animals based on their backgrounds—desert for camels and pasture for cows—rather than on their distinctive features. This reliance can result in misclassifications, such as erroneously identifying a camel in a pasture as a cow. Addressing these spurious correlations is a critical issue across various applications, including medical imaging \cite{oakden2020}, algorithmic fairness \cite{du2021}, and education \cite{perelman2014}.

% para2: spurious correlation을 완화하기 위한 이전 연구들. (bias label이 있는 경우에 previous debiasing studies, 이것의 문제점)
% & spurious correlation을 완화하기 위한 recent 연구들. (bias label이 없는 경우에 previous debiasing studies) (이거 나눌까? 근데 너무 PGD 논문이랑 똑같이쓰면 안됨. 절대로. 근데 이거 이렇게 나누면 또 related work랑 안겹치나? 생각해보고 다른 논문들은 어떻게 썼는지 찾아보기.)
% * 여기서 bias-aligned, bias-conflicting으로 정의한 group에 대해 일반적으로 목표가 무엇인지 (spurious-correlation의 유무에 따라(?)) 명확하게 짚기.
% => models robust to spurious correlations 를 학습하는게 목표. 즉, spurious correlation의 유무에 관계없이, 모든 samples에 대해 좋은 성능을 보이는 model을 얻는것이 목표다.
% bias-aligned/conflicting samples 정의에 대한 내용 (아래 논문들 확인해서 정리해두기.)
% => PGD: Intro 1문단, LWBC Intro 1문단 (We call data with ~)
There have been extensive efforts to reduce the effects of spurious correlations. Numerous studies have initially relied on the assumption that bias attributes are given in the training dataset \cite{groupDRO,data_balancing,DFR,SSA}. However, annotating bias attributes is labor-intensive and expensive, rendering methods dependent on such annotations impractical. Consequently, research has shifted towards developing debiasing techniques that do not require bias labels during training \cite{LfF,DFA,PGD,LCloss,JTT,CNC}. In this line of research, samples with spurious correlations are called \emph{bias-aligned samples}, whereas those lacking such correlations are termed \emph{bias-conflicting samples} \cite{LfF,DFA,LWBC}. In the absence of explicit bias labels, many existing methods focus on identifying bias-conflicting samples and employ strategies such as upweighting or upsampling these samples to counteract the negative impacts of spurious correlations \cite{LfF,JTT,PGD}. Despite their straightforward nature, questions persist regarding the optimal scale for the weight of each sample to effectively reduce dependence on spurious correlations. 

% avoid the reliance on spurious correlations (bias attributes).
% para3: proposed method 제안. 이에 대한 설명 (작동 방법). + 이론 분석도 제공함. (novelty or 이 방법의 당위성)
We thus formulate a debiasing objective for improved robustness against spurious correlations and derive a practical resampling-based algorithm. Specifically, our debiasing objective is built on two groups: the bias-aligned group and the bias-conflicting group, each consisting of bias-aligned and bias-conflicting samples, respectively. This objective encourages the model to perform equally well across both groups. The key insight behind the objective is that models relying on spurious correlations exhibit worse performance on the bias-conflicting group compared to the bias-aligned group, whereas models that do not rely on spurious correlations should exhibit strong performance across both groups. Then, under a simple condition, we derive an objective function as a weighted loss, with weights proportional to the bias-conflicting probability for each training example. Since we consider cases where bias labels are not provided, we use the disagreement probability between the target label and the prediction of an intentionally biased model as a substitute for the bias-conflicting probability. Our proposed approach is simple yet effective in mitigating spurious correlations.

% para4: 내 방법과 이전 방법의 차별점 (novelty), main contribution 요약 (실험 (제안한 방법의 우수성), theoretical analysis) (이 부분은 itemize 이용: notion 확인해서 글 작성하기.)
% Our proposed training objective employs two groups—bias-aligned and bias-conflicting—unlike many prior works that utilize groups defined by combinations of target labels and bias labels. From this objective, we derive DPR, a method that trains models to be robust against spurious correlations without requiring bias labels.

The main contributions of this paper can be summarized as follows:
\begin{itemize}
    \item We present a debiasing objective for mitigating reliance on bias attributes. This objective aims to guide the model to have similar performance over two groups: the bias-aligned group and the bias-conflicting group. This approach differs from previous works, which typically define groups based on combinations of target labels and bias labels.
    
    \item We propose a new method, coined \emph{\textbf{D}isagreement \textbf{P}robability based \textbf{R}esampling for debiasing (\textbf{DPR})}, which is derived from the proposed objective. DPR leverages the disagreement probability between the target label and the prediction of a biased model to determine the weight of each training example.
    
    \item DPR achieves state-of-the-art performance across six benchmarks with spurious correlations, surpassing existing methods that do not use bias labels. Notably, on the bias-conflicting test set of Biased FFHQ (BFFHQ)—which contains only 0.5\% of bias-conflicting samples in the training set—the proposed method significantly improves accuracy by 20.87\% compared to ERM and by 6.2\% compared to the best baseline.
    
    \item We theoretically demonstrate that DPR encourages consistent model performance across both bias-aligned and bias-conflicting groups, thus promoting strong performance across all samples regardless of spurious correlations.
\end{itemize}

\section{Related work}
\label{sc:related}
\paragraph{Debiasing with bias annotations.}
Numerous previous works utilize bias labels for debiasing \cite{groupDRO,LISA,data_balancing,IRM,unshuffle,CSAD}. For example, Group DRO \cite{groupDRO} employs bias labels to define groups and directly enhances worst-group accuracy to mitigate the effect of spurious correlations; meanwhile, LISA \cite{LISA} mixes two samples with the same label but different domains, or with different labels but the same domain, canceling out spurious correlations and learning invariant predictors. These methods have demonstrated their effectiveness across multiple benchmarks with spurious correlations. However, annotating bias labels for each training example is labor-intensive, and obtaining such labels can sometimes be challenging due to privacy concerns as well \cite{zhao2022}. Consequently, in more recent studies, some researchers have utilized only a small set of bias-annotated data to reduce reliance on bias labels \cite{DFR,SSA,CGL}. For example, DFR \cite{DFR} uses a small group-balanced dataset with bias labels to retrain the last layer of the ERM-trained model. 

\paragraph{Debiasing without bias annotations.}
In an effort to eliminate reliance on bias annotations, recent studies predominantly focus on reducing bias without explicit bias labels \cite{LfF,biaswap,DFA,PGD,LCloss,JTT,CNC,uLA}. Given the unavailability of bias labels, these methods commonly employ a two-stage strategy: (1) identifying bias-aligned and bias-conflicting samples, and (2) training the debiased model by leveraging information obtained from (1). The identification of bias-conflicting samples is accomplished through the use of a deliberately biased model, trained either with generalized cross-entropy loss \cite{LfF,biaswap,DFA,PGD,LCloss} or standard cross-entropy loss \cite{JTT,CNC,uLA}. For instance, LfF \cite{LfF} and DFA \cite{DFA} regard samples as bias-conflicting samples if the biased model exhibits higher cross-entropy losses on these samples compared to the debiased model. Subsequent efforts then focus on enhancing classification accuracy for these identified bias-conflicting samples. Numerous studies have proposed diverse debiasing methods \cite{UMIX,MAPLE,EIIL,LWBC}, including those based on contrastive learning \cite{CNC,fighting_fire} and unsupervised clustering \cite{pseudo_attributes,GEORGE}.

Many debiasing methods that do not require bias labels aim to enhance performance on the bias-conflicting group, yet the optimal extent of this enhancement remains uncertain. We contend that a debiased model should exhibit consistent performance across both bias-aligned and bias-conflicting groups. However, to the best of our knowledge, no existing training objective is explicitly designed for this purpose. Therefore, we introduce a novel training objective tailored for this purpose, from which we develop DPR—a debiasing method that does not rely on bias labels.

% referece: Learning Debiased Classifier with Biased Committee, PGD 논문 

\section{Problem formulation}
We consider a multi-class classification problem with a dataset $\mathcal{D} = \{(x_i, y_i)\}_{i=1}^n$, where each $x_i\in\mathcal{X}$ represents an input, and each $y_i\in\mathcal{Y}$ is the corresponding label from $K$ possible classes. These examples are presumed to be sampled from a training distribution $P$. Given a classification model $f_\theta: \mathcal{X} \rightarrow \mathbb{R}^K$ that maps an input to $K$ logits, and a convex loss $\ell:\mathcal{X} \times \mathcal{Y} \rightarrow \mathbb{R}_{\geq 0}$, ERM aims to find a model that minimizes the expected loss $\mathbb{E}_{(x, y) \sim P}[\ell(f_\theta(x), y)]$.  To this end, we typically minimize a surrogate loss $\hat{\mathcal{L}}_{avg}$:
\begin{equation}\label{eq:erm}
    \hat{\mathcal{L}}_{avg} = {1 \over n}\sum_{(x, y) \in \mathcal{D}}{\ell(f_\theta(x), y)}.
\end{equation}
The cross-entropy (CE) loss is commonly used for training classification models. It is defined as $\ell_{\text{CE}}(f_\theta(x), y) = - f_\theta(x)[y] + \log{\sum_{y'}\exp(f_\theta(x)[y'])}$, where $f_\theta(x)[y]$ denotes the logit corresponding to the $y$-th class label.

% correlated/uncorrelated 근거: CNC
We assume there is a spurious correlation presence indicator $b \in \mathcal{B}=\{\text{correlated}, \text{uncorrelated}\}$ in the dataset. As this indicator denotes whether spurious correlations are present within each sample, it can also be considered as a bias-aligned or bias-conflicting group indicator. Similar to the Group DRO setting \cite{groupDRO, DRSL}, we adopt the latent prior probability change assumption \cite{covariate_shift}. With a group indicator $b$, we assume
\begin{equation}\label{eq:cov_shift}
    P(x, y|b) = Q(x, y |b), \quad P(b)\neq Q(b),
\end{equation}
% The training distribution $P$ and test distribution $Q$ are represented as a mixture of group distributions.
% Under this assumption, we can express the training distribution $P(x, y) = \sum_{b \in \mathcal{B}}P(x, y|b)P(b)$ and the test distribution $Q(x, y) = \sum_{b \in \mathcal{B}}P(x, y|b)Q(b)$ 
where $P$ and $Q$ denote the training and test data distributions, respectively. Under this assumption, $P(x, y)$ and $Q(x, y)$ are represented as mixtures of the conditional distributions $P(x, y|b)$. Our goal is to find models that are robust against spurious correlations. Regardless of the presence of spurious correlations within each data example, a model that does not rely on these correlations should exhibit strong performance across all examples. In other words, the debiased model should perform consistently well on both bias-aligned and bias-conflicting groups. To this end, we propose the following training objective to avoid spurious correlations:
\begin{equation}\label{eq:objective}
    \min_{\theta} \max_{b\in\mathcal{B}} \left\{\hat{\mathcal{L}}_{{b}} := {1 \over n_b} \sum_{(x, y, b) \in G_b}\ell(f_\theta(x), y)\right\},
\end{equation}
where $G_b$ is a subset of the training data composed of samples drawn from $P(x, y|b)$ and $n_b$ is the size of $G_b$. The proposed objective aims to minimize the maximum average loss over bias-aligned and bias-conflicting groups, thereby reducing the reliance of classification models on spurious correlations. However, to utilize the above objective, we must know the information about bias attributes. In the next section, we describe a practical method to train the debiased model using \Cref{eq:objective} without bias labels.

\section{DPR: Disagreement Probability based Resampling for debiasing}
% \section{DPR: Disagreement Probability Based Resampling for Debiasing}
\label{method}
We present DPR, a resampling method derived from the proposed objective, which does not require bias annotations during training. First, we provide a walk-through of how the objective in \Cref{eq:objective} can be reformulated as a weighted loss minimization problem. Next, we detail a method for calculating the weight of each training example, along with proxies for both bias-aligned and bias-conflicting groups. Finally, we provide a full description of our algorithm.

\subsection{Problem reformulation} % objective -> resampling method
To utilize the objective in \Cref{eq:objective} without bias annotations, we reformulate this objective as a weighted loss minimization problem. For this purpose, we introduce the following assumption:
% We start the further derivation by introducing the following assumption.
\begin{assumption}
\label{asm:A1}
Let $b_a \in \mathcal{B}$ and $b_c \in \mathcal{B}$ represent the bias-aligned and bias-conflicting groups, respectively. The neural network, parameterized by $\theta$, satisfies that $\hat{\mathcal{L}}_{{b_a}} < \hat{\mathcal{L}}_{{b_c}}$.
\end{assumption}
In \Cref{asm:A1}, we assume that the model, parameterized by $\theta$, exhibits a higher average loss on the bias-conflicting group compared to the bias-aligned group in the training dataset. Under this assumption, the maximum average loss over groups in \Cref{eq:objective} can be expressed as follows:
\begin{align}
    \max_{b \in \mathcal{B}} \hat{\mathcal{L}}_{{b}} &= {1 \over n_{b_c}} \sum_{(x, y, b_c) \in G_{b_c}}\ell(f_\theta(x), y) \label{eq:reformulation1} \\ 
    &= {1 \over n_{b_c}} \sum_{(x, y, b) \in \mathcal{D}}p(b=b_c|x)\ell(f_\theta(x), y) \label{eq:reformulation2} \\
    &= {1 \over n} \sum_{(x, y, b) \in \mathcal{D}}{p(b=b_c|x)\over p(b=b_c)}\ell(f_\theta(x), y). \label{eq:reformulation3}
\end{align}
When \Cref{asm:A1} is satisfied, \Cref{eq:reformulation1} holds. Given that all samples in $G_{b_c}$ have $p(b=b_c|x)$ equal to 1, whereas all samples in $G_{b_a}$ have $p(b=b_c|x)$ equal to 0, \Cref{eq:reformulation2} is derived from \Cref{eq:reformulation1}. As $p(b=b_c)$ is equal to ${n_{b_c} \over n}$, \Cref{eq:reformulation3} is obtained. By denoting ${1 \over n} \cdot {p(b=b_c|x) \over p(b=b_c)}$ as $r(x, y, b)$, we obtain a weighted loss minimization as follows:
\begin{equation}
\label{eq:reweighted}
    \min_{\theta} \sum_{(x, y, b) \in \mathcal{D}}r(x, y, b)\ell(f_\theta(x), y).
\end{equation}
Note that the weight $r(x, y, b)$ can be interpreted as the sampling probability.

% group proxy, disagreement probability: a substitute for the bias-conflicting probability.
\subsection{Sampling probability with group proxy}
\label{howtocomputerx}
In order to train the model using \Cref{eq:reweighted}, it is necessary to compute the sampling probability $r(x, y, b)$ for each training example. However, directly calculating the probabilities $p(b=b_c|x)$ and $p(b=b_c)$ is unfeasible without bias labels. To overcome this limitation, we first introduce proxies for the bias-aligned and bias-conflicting groups. We then derive substitutes for the probabilities $p(b=b_c|x)$ and $p(b=b_c)$ using a biased model.

\paragraph{Group proxy.} We focus on the characteristics of the biased model for the group proxies. Following \citet{LfF}, the biased model $f_\phi$ is trained using ERM with the generalized cross-entropy (GCE) loss \cite{GCE}:
\begin{equation}
\label{eq:gce_loss}
    \ell_{\text{GCE}}(f_\phi(x), y) = {1 - p_\phi(y|x)^q \over q},
\end{equation}
where $p_\phi(y|x)$ represents the probability assigned to the target label $y$ by the neural network after a softmax layer, and $q\in(0, 1]$ is a hyperparameter. The GCE loss amplifies the bias of the model by up-weighting the gradient of the cross-entropy loss for samples with high probability $p_\phi(y|x)$, thereby training the model to rely on spurious correlations. Consequently, the biased model tends to predict correctly for bias-aligned samples and incorrectly for bias-conflicting samples \cite{LfF}. Building on this insight, leveraging the predictions $y_{\text{bias}}$ of the biased model, we employ the agreement between the label and the biased model's prediction (\textit{i.e.}, $y=y_{\text{bias}}$) as a proxy for the bias-aligned group $b_a$, and the disagreement (\textit{i.e.}, $y \neq y_{\text{bias}}$) as a proxy for the bias-conflicting group $b_c$.

\paragraph{Sampling probability.}
We now discuss how to compute the sampling probability $r(x, y, b)$. Using the group proxies mentioned above, we substitute $p(y \neq y_{\text{bias}}| x)$ for $p(b=b_c|x)$. For a given example $(x, y)$, the disagreement probability $p(y \neq y_{\text{bias}}|x)$ can be computed as follows:
\begin{align}
\label{eq:p(y_neq_y_bias_x)}
    p(y \neq y_{\text{bias}}|x) &= \sum_{y_{\text{bias}}}p(y, y_{\text{bias}}|x) - p(y=y_{\text{bias}}|x) = 1 - p_{\text{bias}}(y|x),
\end{align}
where $p_{\text{bias}}(y|x) = {\exp(f_\phi(x)[y]/\tau) \over Z(\phi)}$ is the probability assigned to label $y$ by the biased model, $Z(\phi)=\sum_{y'}\exp(f_\phi(x)[y']/\tau)$ is the partition function, and $\tau$ is a temperature hyperparameter. Note that $p_{\text{bias}}(y|x)$ is used to approximate $p(b=b_a|x)$. It is crucial for the biased model to accurately capture the spurious correlation structure; hence, the probabilities of the biased model should be appropriately calibrated \cite{oncalibration,whendoes,onthestatistical}. We also substitute $p(y \neq y_{\text{bias}})$ for $p(b=b_c)$. This probability can be estimated using all the training data, as in prior works \cite{LCloss,uLA}:
\begin{equation}
\label{eq:p(y_neq_y_bias)}
    p(y \neq y_{\text{bias}}) \approx {1 \over n} \sum_{(x, y, b) \in \mathcal{D}}p(y \neq y_{\text{bias}}|x) = {1 \over n} \sum_{(x, y, b) \in \mathcal{D}} \left(1 - p_{\text{bias}}(y|x)\right).
\end{equation}
With \Cref{eq:p(y_neq_y_bias_x),eq:p(y_neq_y_bias)}, we compute 
\begin{equation}
\label{eq:r_hat}
    \hat{r}(x, y) = {1 \over n} \cdot {p(y \neq y_{\text{bias}} | x) \over p(y \neq y_{\text{bias}})} = {1 - p_{\text{bias}}(y|x) \over \sum_{(x, y, b) \in \mathcal{D}}\left(1 - p_{\text{bias}}(y|x)\right)}.
\end{equation}
Instead of $r(x, y, b)$, we use $\hat{r}(x, y)$ as the sampling probability. We are now ready to train the debiased model.

\begin{algorithm}[t]
\caption{Disagreement Probability based Resampling for debiasing (DPR)}\label{algorithm}
\KwData{training set $\mathcal{D}$, biased model $f_\phi$, debiased model $f_\theta$, learning rate $\eta$, the total number of iterations $T_b$ and $T_d$, calibration parameter $\tau$, GCE parameter $q$}
% \KwResult{Write here the expected result}
\BlankLine
\tcc{Train the biased model}
\For{$t=1$ \KwTo $T_b$}{
    Sample a mini-batch $\{(x, y)\}$ from $\mathcal{D}$\\
    Update $\phi$ by training on a mini-batch using \Cref{eq:gce_loss}
}
\BlankLine
\tcc{Compute the sampling probability}
Compute $\hat{r}(x, y)$ for all $(x, y) \in \mathcal{D}$ following \Cref{eq:r_hat} \\
\BlankLine
\tcc{To satisfy \Cref{asm:A1}}
Initialize the debiased model $f_\theta$ with the biased model $f_\phi$ \\
\BlankLine
\tcc{Train the debiased model}
\For{$t=1$ \KwTo $T_d$}{
    Sample a mini-batch $\{(x, y)\}$ from $\mathcal{D}$ according to $\hat{r}(x, y)$ \\
    Update $\theta$ by training on a mini-batch using cross-entropy loss
}
\end{algorithm}

\subsection{Training algorithm} % 전체 알고리즘 설명 + assumption 만족을 위한 과정 + data augmentation + algorithm 작성
\label{sc:algo}
We outline the entire training process for our method as follows. First, we train the biased model $f_\phi$. Next, we compute the sampling probability $\hat{r}(x, y)$ using the pretrained biased model. Before proceeding to train the debiased model $f_\theta$, it is worth noting that \Cref{eq:reweighted} is derived under \Cref{asm:A1}. Thus, it is essential to fulfill \Cref{asm:A1} when training the debiased model using \Cref{eq:reweighted}. We leverage the characteristics of the biased model for this purpose. The biased model typically exhibits higher loss on bias-conflicting samples and lower loss on bias-aligned samples, thereby fulfilling \Cref{asm:A1}. Consequently, we initialize the model $f_\theta$ with the biased model $f_\phi$ and then train $f_\theta$ using training examples sampled with the probability $\hat{r}(x, y)$. We also employ data augmentation to enhance the diversity of bias-conflicting samples. Simply oversampling these samples without enhancing their diversity does not effectively mitigate bias \cite{DFA}. Therefore, we enhance the diversity of bias-conflicting samples through data augmentation techniques such as random color jitter and random rotation. The complete training procedure is outlined in \Cref{algorithm}. 

\section{Theoretical analysis}
% 처음 도입부터 아예, "minimizing our loss 는 bias-aligned/bias-conflicting groups의 performance를 둘 다 향상시킨다 while maintaining similar performance across them." 이라는 주장을 -> "we theoretically show that our proposed debiasing objective minimizes the loss of both bias-aligned and bias-conflicting groups while encouraging to maintain similar loss across them. "
In this section, we theoretically demonstrate that DPR minimizes losses for both bias-aligned and bias-conflicting groups while reducing the disparity between their losses. All proofs are deferred to \Cref{proof}. Let $\mathcal{L}_{avg}$ be the expected average loss:
\begin{equation}
    \mathcal{L}_{avg} := \mathbb{E}_{(x, y) \sim P}[\ell(f_\theta(x), y)].
\end{equation}
Let $\mathcal{L}_b$ be the average loss of group $b$:
\begin{equation}
    \mathcal{L}_b := \mathbb{E}_{(x, y) \sim P_b}[\ell(f_\theta(x), y)], 
\end{equation}
where $P_b=P(x, y|b)$ denotes the training distribution conditioned on $b$, for any $b \in \mathcal{B}$. In this setting, we derive the following inequality for the loss gap between the bias-aligned and bias-conflicting groups.
\begin{theorem}
\label{theorem1}
Suppose that the loss function $\ell(f_\theta(x), y)$ is upper-bounded by a constant $C>0$. Given two distinct groups $b_a \in \mathcal{B}$ and $b_c \in \mathcal{B}$ such that $b_a \neq b_c$, the following inequality holds with probability at least $1-\delta$, for any $\delta > 0$:
\begin{equation}
\label{eq:thr1}
    |\mathcal{L}_{b_a} - \mathcal{L}_{b_c}| \leq 2 \cdot \max_{b\in\mathcal{B}}\hat{\mathcal{L}}_{{b}} + C \cdot \max_{b \in \mathcal{B}} \sqrt{{8\log{(|\mathcal{B}|/\delta)} \over n_{b}}}. 
    % Hoeffding's inequality: https://en.wikipedia.org/wiki/Hoeffding%27s_inequality
    % googling: generalization bound and hoeffding inequality
    % 참고: CNC eq(13)에서 max 부분은 그냥 위 수식에서 root가 가장 큰 group g에 대해 2배하면 나오는 값인 것 같음.
    % 참고: https://engineering.purdue.edu/ChanGroup/ECE595/files/chapter4.pdf <- 여기 Theorem 2 (M과 |G|의 연관성), 3. 중심으로 수식 확인.
\end{equation}
\end{theorem}
\Cref{theorem1} specifies that the upper bound on the disparity between losses for bias-aligned and bias-conflicting groups is determined by the maximum average loss over groups and a term dependent on the size of the smaller group. Additionally, we derive an inequality associated with the expected average loss.
\begin{theorem}
\label{theorem2}
In the same setting as \Cref{theorem1}, the expected average loss is bounded above with probability at least $1-\delta$:
\begin{equation}
\label{eq:thr3}
    \mathcal{L}_{\text{avg}} \leq \max_{b\in\mathcal{B}}\hat{\mathcal{L}}_{{b}} + C \cdot \sqrt{2\log(1/\delta) \over n}.
\end{equation}
\end{theorem}
According to \Cref{theorem1,theorem2}, our proposed training objective not only closes the loss gap between bias-aligned and bias-conflicting groups but also reduces the expected average loss. However, in \Cref{eq:thr1}, there remains a loss gap due to a term inversely related to the square root of the size of the smaller group. Note that DPR is a resampling method derived from the proposed objective when \Cref{asm:A1} is fulfilled, and it identifies and upsamples bias-conflicting samples. Thus, it efficiently minimizes both terms of the upper bound described in \Cref{eq:thr1}. Given that $\mathcal{L}_{avg}$ is expressed as a weighted sum of $\mathcal{L}_{b_a}$ and $\mathcal{L}_{b_c}$, these theorems indicate that DPR enhances performance across both bias-aligned and bias-conflicting groups while reducing the performance gap between them.
% Note: Intro에서 spurious correlations 는 majority of the training examples 에 존재한다고 설명했음. 그리고 spurious correlations가 있는 samples를 bias-aligned, such correlations가 없는 samples를 bias-conflicting samples라고 한다고 했음. 그렇다면, 이로부터 # of bias-aligned samples > # of bias-conflicting samples 가 유추될 것이다.

\section{Experiments}
\label{sc:exp}
\subsection{Datasets}
% CMNIST, MBMNIST, BAR, bFFHQ, CelebA, CivilComments
We evaluate the debiasing performance of DPR using six benchmark datasets that exhibit spurious correlations. Colored MNIST and Multi-bias MNIST are synthetic datasets designed under the premise that models learn bias attributes first. Conversely, BAR, BFFHQ, CelebA, and CivilComments-WILDS are real-world datasets where inherent biases degrade model performance. We follow the evaluation protocols of previous studies \cite{PGD,DFA,JTT} and provide a detailed description of these datasets in \Cref{apdx:dataset}.

\paragraph{Colored MNIST.}
Colored MNIST (C-MNIST) is a synthetic dataset designed for digit classification, comprising ten digits, each spuriously correlated with a specific color. Following the protocols in \citet{PGD}, we set the ratios of bias-conflicting samples, denoted as $\rho$, at 0.5\%, 1\%, and 5\% for the training set, and 90\% for the unbiased test set. We report the accuracy on this unbiased test set.

\paragraph{Multi-bias MNIST.}
Multi-bias MNIST (MB-MNIST) \cite{PGD} is a synthetic dataset designed to incorporate more complex patterns compared to C-MNIST and biased-MNIST \cite{biased_mnist}. MB-MNIST comprises eight attributes: digit \cite{mnist}, alphabet \cite{emnist}, fashion object \cite{fmnist}, Japanese character \cite{japanese}, digit color, alphabet color, fashion object color, and Japanese character color. The digit shape is the target attribute, while the other seven serve as bias attributes. In MB-MNIST, bias is introduced by aligning the digit with each of the other seven attributes, each with a probability of $1-\rho$. In our experiments, $\rho$ is set to 10\%, 20\%, and 30\% for the training set and 90\% for the unbiased test set, as in \citet{PGD}. We report the accuracy on this unbiased test set.

\paragraph{Biased action recognition.}
The biased action recognition (BAR) dataset \cite{LfF}, designed for action classification, comprises six action classes such as climbing and fishing, each spuriously correlated with a specific place. The training set of BAR contains only bias-aligned samples, while the test set consists solely of bias-conflicting samples. We report the accuracy on this bias-conflicting test set.

\paragraph{Biased FFHQ.}
Biased FFHQ (BFFHQ) is a real-world facial dataset, which has age (young or old) as a label and gender (male or female) as a bias attribute. Predominantly, females are young and males are old in this dataset. We use a bias-conflicting ratio of $\rho=0.5\%$ in the training set and report accuracies on both an unbiased test set with $\rho=50\%$ \cite{PGD} and a bias-conflicting test set with $\rho=100\%$ \cite{DFA}.

\paragraph{CelebA.}
CelebA \cite{CelebA} is a dataset for facial classification. The goal is to classify celebrity hair color as blond or non-blond, which is spuriously correlated with gender. Notably, only a few blond-haired celebrities are male. Following prior studies \cite{CNC,groupDRO}, we report both average and worst-group accuracies on the test set, where groups are defined as combinations of class labels and bias labels.

\paragraph{CivilComments-WILDS.}
CivilComments-WILDS \cite{civilcomments_borkan,wilds} is a text classification dataset aimed at identifying whether online comments are toxic or non-toxic. The label is spuriously correlated with demographic identities such as gender, race, and religion. Following previous works \cite{JTT,wilds,CNC}, we report both average and worst-group accuracies on the test set, where groups are defined as combinations of class labels and bias labels.

\subsection{Experimental setup}
\paragraph{Baselines.}
% 6개 쓰는게 적당할 것 같다. 
We compare our method with six baselines on various benchmarks: ERM, JTT \cite{JTT}, DFA \cite{DFA}, CNC \cite{CNC}, PGD \cite{PGD}, and LC \cite{LCloss}. ERM denotes conventional training without any considerations for debiasing, while the others are debiasing methods that do not require bias annotations during training.

\paragraph{Implementation details.}
% image processing (augmentation), architecture, hyperparameters
For all datasets except CelebA and CivilComments-WILDS, we follow the experimental settings of \citet{PGD}. Specifically, for CMNIST and MB-MNIST, we employ two distinct types of simple CNN models, respectively. For BAR and BFFHQ, we utilize a ResNet18 \cite{resnet} pretrained on ImageNet \cite{imagenet}. In the case of CelebA, we use a pretrained ResNet50, following the experimental settings of \citet{CNC}. For CivilComments-WILDS, we deploy a pretrained BERT model and follow the experimental setup detailed in \citet{JTT}. Moreover, we apply data augmentation techniques—including random resize crop, random color jitter, and random rotation—for all vision datasets except CelebA, as discussed in \Cref{sc:algo}. Further details on model architectures, hyperparameters, and image processing are provided in \Cref{apdx:impdetail}.

% Following \cite{groupDRO}, we report all results with early stopping with respect to worst-group validation accuracy.

% As proposed in \cite{groupDRO}, we report all results with early stopping based on the validation accuracy. For all datasets except CelebA and CivilComments-WILDS, we select the model 

% we perform early stopping with accuracy on validation set and report mean and standard deviation across 3 runs.

\subsection{Experimental results and analysis}
\paragraph{Classification accuracy.} 
\Cref{table:synthetic} presents the average accuracies on unbiased test sets for C-MNIST and MB-MNIST. DPR consistently outperforms or matches the performance of other baselines across all experiments with varying bias-conflicting ratios. Notably, for the MB-MNIST dataset with a ratio of 30\%, DPR achieves an unbiased test accuracy of 94.04\%, outperforming PGD by 3.28\%. Even for MB-MNIST with a ratio of 10\%, where all baselines except PGD fail to achieve higher accuracy, DPR exhibits the highest accuracy of 62.21\%. In the more complex setting of MB-MNIST, compared to C-MNIST, the effectiveness of DPR is even more pronounced. The superiority of our method is further demonstrated on real-world benchmarks, as shown in \Cref{table:realworld}. Our method consistently achieves the best performance for each real-world dataset. Specifically, for the BFFHQ dataset, DPR achieves an accuracy of 87.57\% on unbiased test sets, which is 3.37\% higher than PGD, and an accuracy of 76.80\% on bias-conflicting test sets, which is 6.20\% higher than LC. For the CelebA and CivilComments-WILDS datasets, our method achieves the highest worst-group accuracy compared to other baselines, with groups defined as the combination of the target and bias labels. We especially highlight the results for the BFFHQ benchmark, where our method improves accuracy on both unbiased and bias-conflicting test sets. This result supports our claim that DPR enhances performance on both bias-aligned and bias-conflicting groups.
% \begin{table}[thb!]
\begin{table}[t]
\centering
\caption{Average accuracies and standard deviations over three trials on two synthetic image datasets, C-MNIST and MB-MNIST, under varying ratios (\%) of bias-conflicting samples. Except for LC, the results of baselines reported in Ahn et al. \cite{PGD} are provided. The best performances are highlighted in \textbf{bold}.}
\label{table:synthetic}
\resizebox{0.78\textwidth}{!}{%
\begin{tabular}{l|cccccc}
\toprule[1pt]
           & \multicolumn{3}{c}{C-MNIST}                                           & \multicolumn{3}{c}{MB-MNIST}                                          \\ \cmidrule(lr){2-4} \cmidrule(lr){5-7}
Ratio (\%) & 0.5                   & 1                     & 5                     & 10                    & 20                    & 30                    \\ \midrule
ERM        & 60.94 (0.97)          & 79.13 (0.73)          & 95.12 (0.24)          & 25.23 (1.16)          & 62.06 (2.45)          & 87.61 (1.60)          \\
JTT        & 85.84 (1.32)          & 95.07 (3.42)          & 96.56 (1.23)          & 25.34 (1.45)          & 68.02 (3.23)          & 85.44 (3.44)          \\
DFA        & 94.56 (0.57)          & 96.87 (0.64)          & 98.35 (0.20)          & 25.75 (5.38)          & 61.62 (2.60)          & 88.36 (2.06)          \\
PGD        & 96.88 (0.28)          & 98.35 (0.12)          & \textbf{98.62 (0.14)} & 61.38 (4.41)          & 89.09 (0.97)          & 90.76 (1.84)          \\
LC         & 97.25 (0.21)          & 97.34 (0.16)          & 97.44 (0.37)          & 25.86 (8.68)          & 71.23 (1.71)          & 89.57 (2.50)          \\
DPR (Ours)       & \textbf{97.52 (0.33)} & \textbf{98.40 (0.03)} & \textbf{98.62 (0.12)} & \textbf{62.21 (4.02)} & \textbf{89.11 (1.65)} & \textbf{94.04 (0.26)} \\ \bottomrule[1pt]
\end{tabular}%
}
\vspace{-4.0mm}
\end{table}
% \begin{table}[thb!]
\begin{table}[t]
\centering
% TODO: 아래 Conflicting, Unbiased, Average, Worst <- 이거를 더 깔끔하게 다른걸로 바꿀 수 없을까? emph? ``''? 다른 논문들은 어떻게 했는지 확인해보고 바꿔야될 것 같으면 바꾸기.
\caption{Classification accuracies (\%) and standard deviations over three trials on four real-world datasets. Conflicting refers to accuracy on bias-conflicting test sets, while Unbiased indicates accuracy on unbiased test sets. Average and Worst denote average test accuracy and worst-group accuracy, respectively. Results of all baselines except LC are taken from Ahn et al. \cite{PGD}, with the exception of bias-conflicting accuracy on BFFHQ. The best performances are highlighted in \textbf{bold}.}
\label{table:realworld}
\resizebox{0.85\textwidth}{!}{%
\begin{tabular}{l|clclc|cccc}
\toprule[1pt]
\multicolumn{1}{c|}{} & \multicolumn{2}{c}{BAR}                   & \multicolumn{3}{c|}{BFFHQ}                                        & \multicolumn{2}{c}{CelebA}       & \multicolumn{2}{c}{CivilComments-WILDS} \\ \cmidrule(lr){2-3} \cmidrule(lr){4-6} \cmidrule(lr){7-8} \cmidrule(lr){9-10}
Accuracy (\%)         & \multicolumn{2}{c}{Conflicting}           & \multicolumn{2}{c}{Unbiased}              & Conflicting           & Average    & Worst               & Average        & Worst                  \\ \midrule
ERM                   & \multicolumn{2}{c}{63.15 (1.06)}          & \multicolumn{2}{c}{77.77 (0.45)}          & 55.93 (0.64)          & 94.9 (0.3) & 47.7 (2.1)          & 92.1 (0.4)     & 58.6 (1.7)             \\
JTT                   & \multicolumn{2}{c}{63.62 (1.33)}          & \multicolumn{2}{c}{77.93 (2.16)}          & 56.13 (0.83)          & 88.1 (0.3) & 81.5 (1.7)          & 91.1 (-)       & 69.3 (-)               \\
DFA                   & \multicolumn{2}{c}{64.70 (2.06)}          & \multicolumn{2}{c}{82.77 (1.40)}          & 66.00 (2.00)          & -          & -                   & -              & -                      \\
CNC                   & \multicolumn{2}{c}{-}                     & \multicolumn{2}{c}{-}                     & -                     & 89.9 (0.5) & 88.8 (0.9)          & 81.7 (0.5)     & 68.9 (2.1)             \\
PGD                   & \multicolumn{2}{c}{65.39 (0.47)}          & \multicolumn{2}{c}{84.20 (1.15)}          & 70.07 (2.00)          & 88.6 (-)   & 88.8 (-)            & 92.1 (-)       & 70.6 (-)               \\
LC                    & \multicolumn{2}{c}{63.45 (2.14)}          & \multicolumn{2}{c}{83.97 (0.83)}          & 70.60 (0.60)          & -          & 88.1 (0.8)          & -              & 70.3 (1.2)             \\
DPR (Ours)                  & \multicolumn{2}{c}{\textbf{66.11 (3.29)}} & \multicolumn{2}{c}{\textbf{87.57 (1.22)}} & \textbf{76.80 (2.51)} & 90.7 (0.6) & \textbf{88.9 (0.6)} & 82.9 (0.7)     & \textbf{70.9 (1.7)}    \\ \bottomrule[1pt]
\end{tabular}%
}
\vspace{-3.0mm}
\end{table}

% \begin{figure}[htb!]
\begin{figure}[t!]
\vspace{-2.0mm}
\centering
\subfigure[C-MNIST (0.5\%)]{\label{fig:group-est-0.5p}\includegraphics[width=40mm]{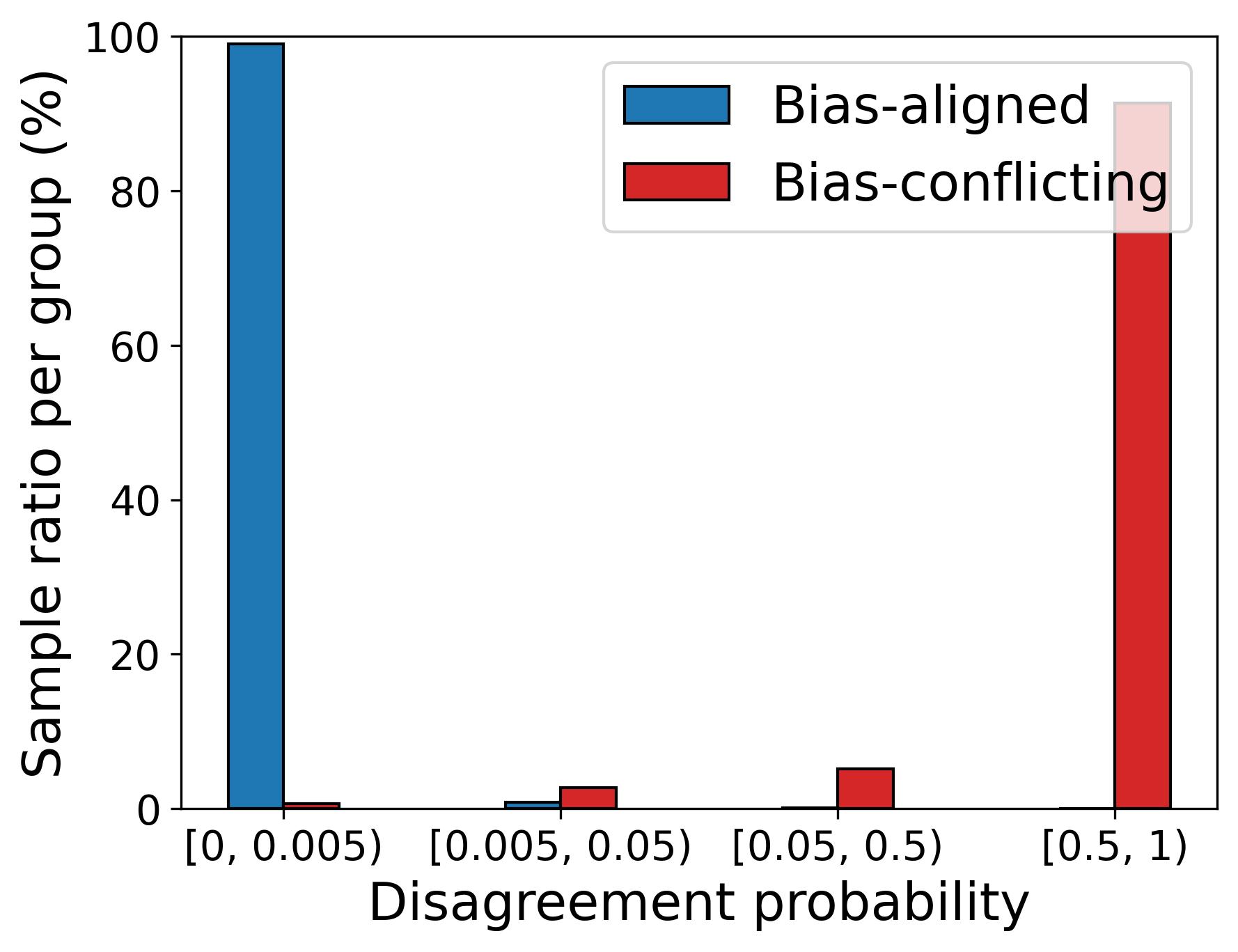}}
\subfigure[C-MNIST (1\%)]{\label{fig:group-est-1p}\includegraphics[width=40mm]{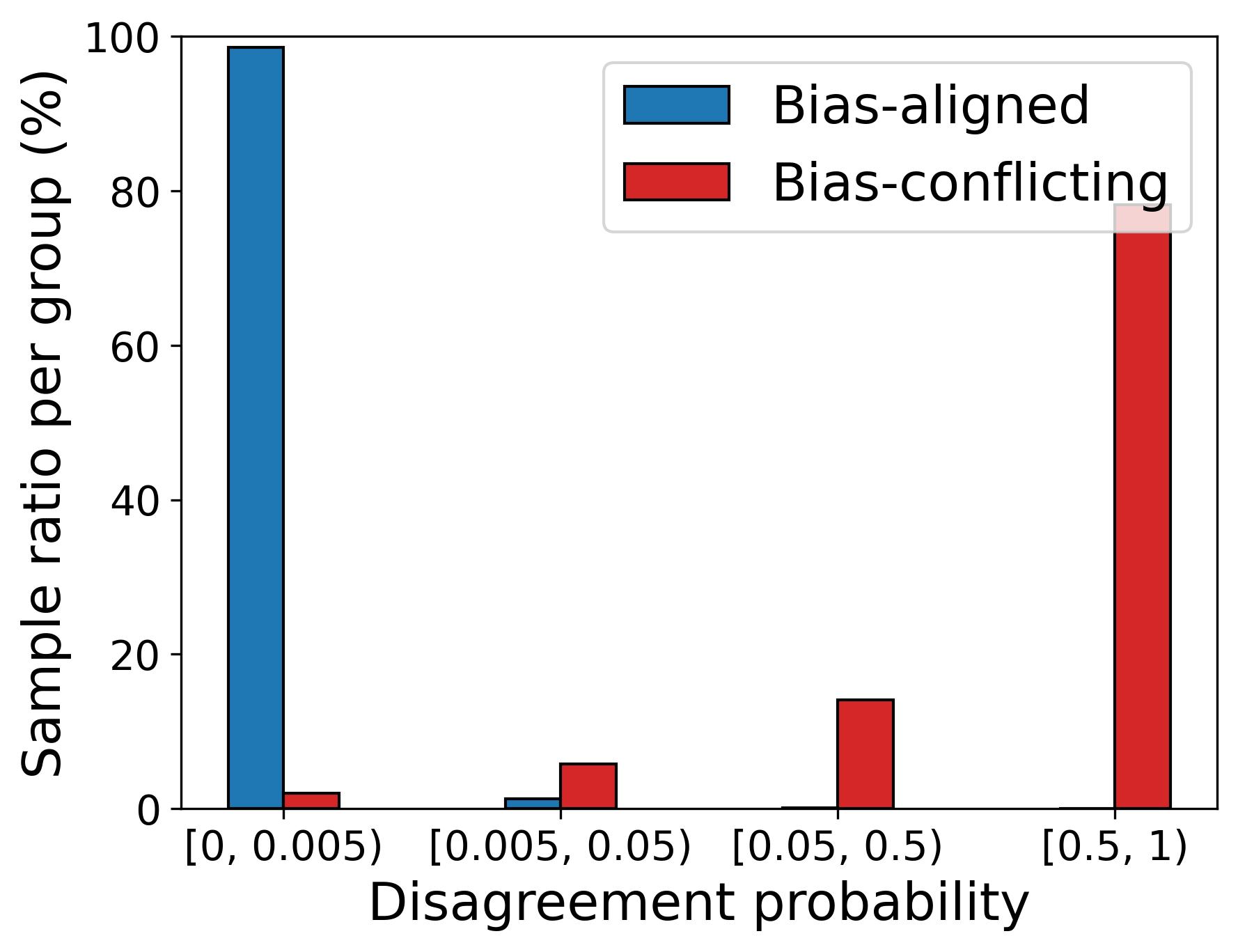}}
\subfigure[C-MNIST (5\%)]{\label{fig:group-est-5p}\includegraphics[width=40mm]{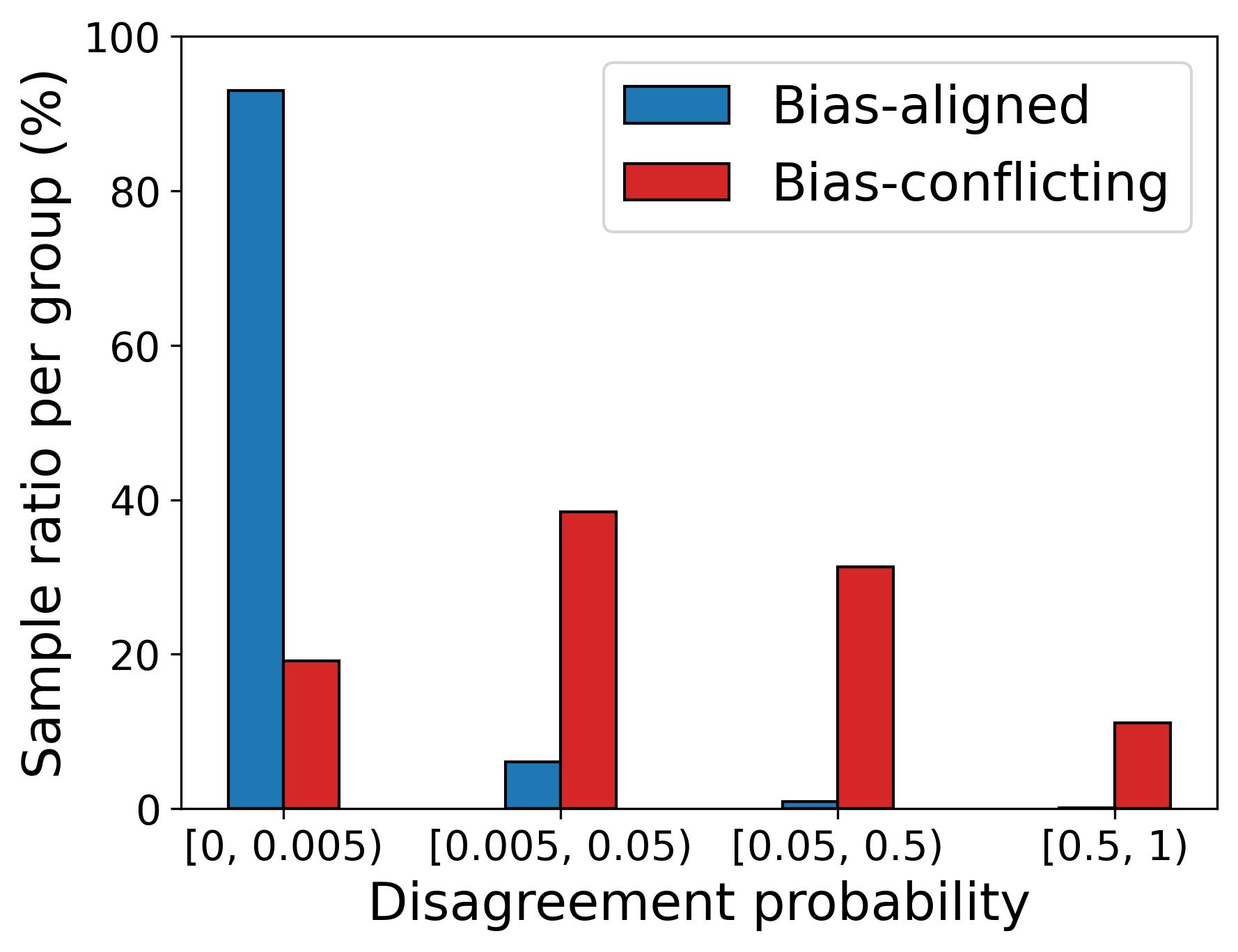}}
\caption{Distributions of disagreement probabilities for each sample within bias-aligned and bias-conflicting groups.}
\label{fig:group-est}
\vspace{-6.0mm}
\end{figure}
\paragraph{Identifying group via disagreement probability.}
To discern whether each sample belongs to the bias-aligned or bias-conflicting group, we check if the target label disagrees with the prediction of the biased model and use its probability as an up-weight. To evaluate whether the disagreement probability between the target label and the prediction of the biased model distinguishes bias-conflicting samples from bias-aligned samples effectively, we plot the distributions of disagreement probabilities $p(y \neq y_{\text{bias}}|x)$ for each bias-aligned and bias-conflicting sample. The experiment is conducted on the C-MNIST dataset. As illustrated in \Cref{fig:group-est}, bias-aligned samples generally exhibit smaller disagreement probabilities compared to bias-conflicting samples. This result demonstrates that disagreement probability effectively differentiates bias-aligned and bias-conflicting samples. Moreover, the relatively high disagreement probability associated with bias-conflicting samples enables DPR to effectively identify and upsample bias-conflicting samples, suggesting that the disagreement probability $p(y \neq y_{\text{bias}}|x)$ serves as an efficient substitute for the bias-conflicting probability $p(b = b_c|x)$.

% 아래 문단의 목적: y!=y_bias 는 b=b_c에 대한 타당한 estimation? proxy? 인가? 그리고 그에 대한 probability도 타당한 proxy 인가? 를 알아보기 위함.
% (p(y!=y_bias|x) 가 p(b=b_c|x) 를 완벽하게 추정하는건 아니더라도, 아래 그림을 통해 말할 수 있는 것: 1. bias-aligned/bias-conflicting samples를 효과적으로 구분하며, 2. bias-conflicting samples 에는 더 큰 probability를 할당하고, aligned samples에는 상대적으로 작은 probability를 할당하여 원래의 목표 "resampling method가 bias-conflicting samples를 타겟해서 upsampling 할 수 있도록" 를 어느정도 달성할 수 있도록 만든다. 따라서 이는 efficient proxy 로 볼 수 있다? 

% \paragraph{Impact of model initialization on loss.}
% \paragraph{Assumption satisfaction.}
% \paragraph{Impact of Model type on Loss.}
% \paragraph{Influence of model initialization.}
% \paragraph{Loss analysis.}
% \paragraph{Loss gap of diverse models between two groups.}

% \paragraph{Validity of debiased model initialization.}
% The Loss Discrepancy Across Groups
% \paragraph{Loss of different models over groups.}
% \paragraph{Loss disparities of different models.}
% \paragraph{Loss of various models on bias-aligned and bias-conflicting groups.}
\paragraph{Group losses of different models.}
\begin{wrapfigure}{h}{0.33\columnwidth}
    \vspace{-5.0mm}
    \centerline{\includegraphics[width=0.33\columnwidth]{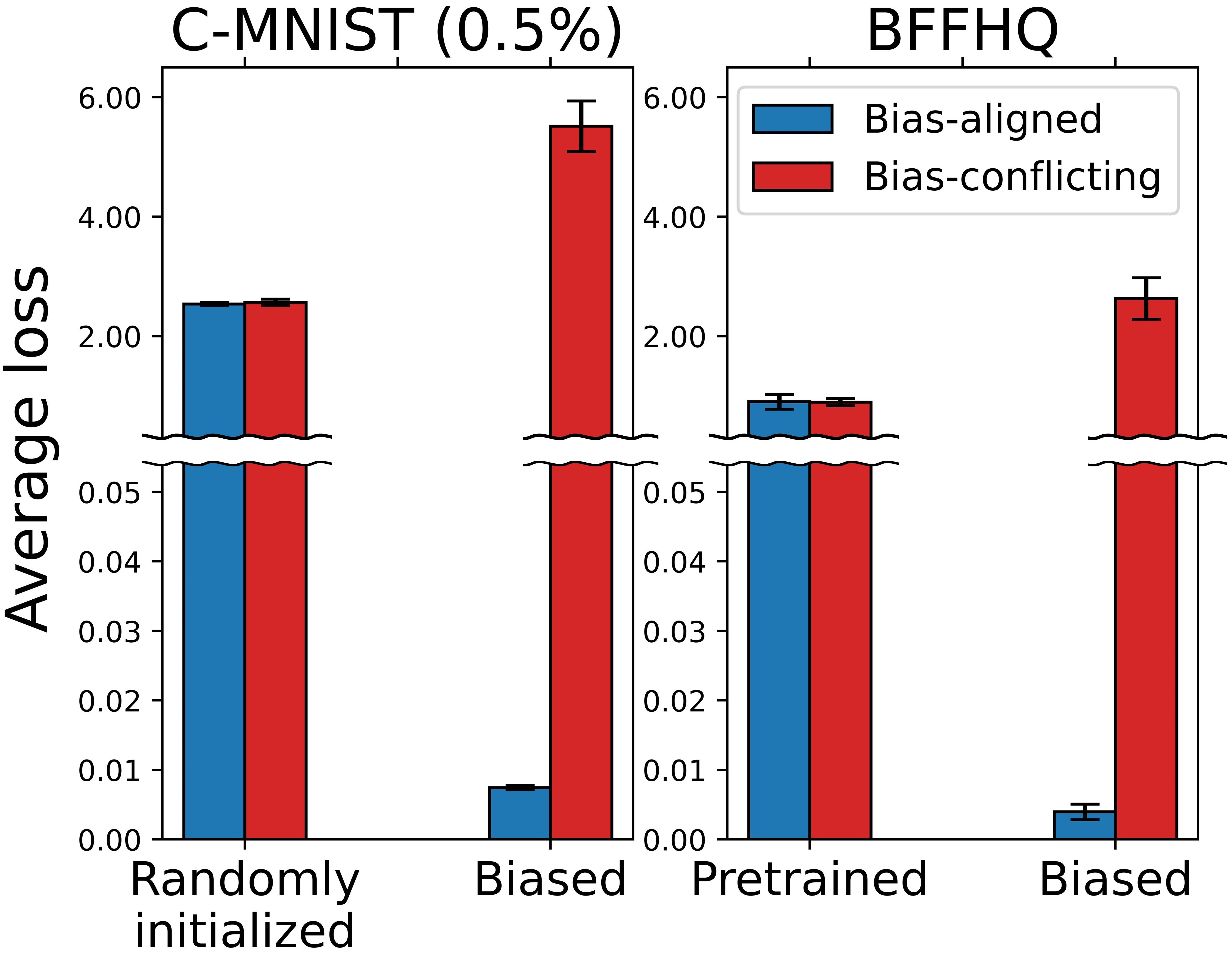}}
    \caption{Average loss of randomly initialized, pretrained, and biased models on bias-aligned and bias-conflicting groups. The error bars represent the standard deviations over three trials.}
    \label{fig:asm-loss}
    \vspace{-5.0mm}
\end{wrapfigure}
As stated earlier in \Cref{sc:algo}, our debiasing objective in \Cref{eq:reweighted} should be used only when \Cref{asm:A1} is satisfied. A natural next question is: which type of model satisfies this assumption? We test three types of models: a randomly initialized model, a model pretrained on ImageNet, and a biased model. Specifically, we examine the average loss of a randomly initialized model on the bias-aligned and bias-conflicting groups of C-MNIST with a bias-conflicting ratio of 0.5\%. Additionally, we examine the average loss of a pretrained model on the bias-aligned and bias-conflicting groups of BFFHQ. We also investigate the average loss of a biased model on the bias-aligned and bias-conflicting groups of C-MNIST and BFFHQ. The results, shown in \Cref{fig:asm-loss}, indicate that both the randomly initialized and pretrained models have similar average losses on the bias-aligned and bias-conflicting groups. In contrast, the biased model has a higher average loss on the bias-conflicting group compared to the bias-aligned group for both C-MNIST and BFFHQ. This result supports the validity of initializing the debiased model with the biased model.

\paragraph{Ablation study.}
We conduct an ablation study on model initialization, GCE, and data augmentation. Model initialization refers to initializing the debiased model with the biased model. The evaluation is performed on C-MNIST with bias-conflicting ratios of 0.5\% and 5\%, as well as BFFHQ. \Cref{table:ablt} demonstrates the importance of each component. Comparing the first and second rows, we observe that GCE and data augmentation bring small improvements when initialization is not used. However, from rows 3-5, we observe that GCE and augmentation bring significant improvements when initialization is utilized. For BFFHQ, introducing both GCE and augmentation significantly improves the average accuracies on unbiased and bias-conflicting sets by 9.07\% and 19.33\%, respectively. Furthermore, we observe the performance gap between using and not using initialization. These results suggest that model initialization is crucial and that GCE and augmentation are also important when initialization is introduced. The best performances are consistently achieved when all components are utilized.
% \begin{table}[htb!]
\begin{table}[t]
\centering
\caption{Ablation studies of the proposed method on the C-MNIST and BFFHQ datasets. We report the average test accuracies and standard deviations over three trials on unbiased and bias-conflicting test sets. A checkmark (\cmark) indicates the case where each component of the proposed method is applied. The best performances are highlighted in \textbf{bold}.}
\label{table:ablt}
\resizebox{0.82\textwidth}{!}{%
\begin{tabular}{ccccccc}
\toprule[1pt]
               &       &              & C-MNIST (0.5\%) & C-MNIST (5\%) & \multicolumn{2}{c}{BFFHQ}   \\ \cmidrule(lr){4-4} \cmidrule(lr){5-5} \cmidrule(lr){6-7}
Initialization & GCE   & Augmentation & Unbiased        & Unbiased      & Unbiased     & Conflicting  \\ \midrule
\xmark          & \xmark & \xmark        & 66.13 (0.51)    & 95.32 (0.42)  & 77.07 (1.16) & 54.60 (2.43) \\
\xmark          & \cmark & \cmark        & 66.47 (1.74)    & 95.21 (0.14)  & 77.60 (0.79) & 55.60 (1.64) \\
\cmark          & \xmark & \xmark        & 89.06 (0.62)    & 97.36 (0.29)  & 78.50 (0.50) & 57.47 (0.90) \\
\cmark          & \cmark & \xmark        & 95.94 (0.45)    & 97.66 (0.17)  & 80.93 (1.33) & 63.40 (3.67) \\
\cmark          & \cmark & \cmark        & \textbf{97.52 (0.33)}    & \textbf{98.62 (0.12)}  & \textbf{87.57 (1.22)} & \textbf{76.80 (2.51)} \\ \bottomrule[1pt]
\end{tabular}%
}
\vspace{-4.0mm}
\end{table}

% reference: The Effects of Regularization and Data Augmentation are Class Dependent
\section{Conclusions and Limitations}
We present DPR, a resampling method that leverages disagreement probability to identify and upsample bias-conflicting samples. This method is derived from a novel training objective designed to achieve consistent performance across both bias-aligned and bias-conflicting groups. It does not rely on bias annotations and has demonstrated state-of-the-art performance across spurious correlation benchmarks. However, DPR has certain limitations: its effectiveness depends on how well the biased model captures the spurious correlation structure, as it uses the predictions of this model to infer the group to which each training sample belongs. Moreover, DPR employs a two-stage approach that complicates the training process and introduces additional hyperparameters. Despite these limitations, DPR's simple implementation and strong performance, supported by theoretical analysis illustrating its ability to reduce loss disparity between groups and minimize average loss, underscore its usefulness in mitigating reliance on spurious correlations.

\section*{Acknowledgements}
This work is in part supported by the National Research Foundation of Korea (NRF, RS-2024-00451435(25\%), RS-2024-00413957(25\%)), Institute of Information \& communications Technology Planning \& Evaluation (IITP, 2021-0-01059(15\%), 2021-0-00106(20\%), 2021-0-00180(15\%)) grant funded by the Ministry of Science and ICT (MSIT), a grant-in-aid of HANHWA SYSTEMS, Institute of New Media and Communications(INMAC), and the BK21 FOUR program of the Education and Research Program for Future ICT Pioneers, Seoul National University in 2024.

% \section*{References}

\bibliography{ref}
\bibliographystyle{plainnat}

% References follow the acknowledgments in the camera-ready paper. Use unnumbered first-level heading for
% the references. Any choice of citation style is acceptable as long as you are
% consistent. It is permissible to reduce the font size to \verb+small+ (9 point)
% when listing the references.
% Note that the Reference section does not count towards the page limit.
% \medskip

% {
% \small

% [1] Alexander, J.A.\ \& Mozer, M.C.\ (1995) Template-based algorithms for
% connectionist rule extraction. In G.\ Tesauro, D.S.\ Touretzky and T.K.\ Leen
% (eds.), {\it Advances in Neural Information Processing Systems 7},
% pp.\ 609--616. Cambridge, MA: MIT Press.

% [2] Bower, J.M.\ \& Beeman, D.\ (1995) {\it The Book of GENESIS: Exploring
%   Realistic Neural Models with the GEneral NEural SImulation System.}  New York:
% TELOS/Springer--Verlag.

% [3] Hasselmo, M.E., Schnell, E.\ \& Barkai, E.\ (1995) Dynamics of learning and
% recall at excitatory recurrent synapses and cholinergic modulation in rat
% hippocampal region CA3. {\it Journal of Neuroscience} {\bf 15}(7):5249-5262.
% }

%%%%%%%%%%%%%%%%%%%%%%%%%%%%%%%%%%%%%%%%%%%%%%%%%%%%%%%%%%%%%%%%%%%%%%%%%%%%%%%%%%%%%%%%%%%%%%%%%%%%%%%%%%%%%%%%%%%%%%%%

\newpage
\appendix

\section*{Appendix}

\section{Missing proofs from Section 5}
\label{proof}
\renewcommand*{\proofname}{Proof of \Cref{theorem1}}
\begin{proof}
Consider the two groups $b_a\in\mathcal{B}$ and $b_c\in\mathcal{B}$ such that $b_a \neq b_c$. Let $d(b_a, b_c)$ be the difference in expected losses between the group $b_a$ and $b_c$:
% Let $n_{b_a}$ and $n_{b_c}$ be the number of examples belonging to groups $b_a$ and $b_c$ in the training dataset, respectively.
\begin{equation}
    d(b_a, b_c) = \left| \mathcal{L}_{b_a} - \mathcal{L}_{b_c} \right|.
    % d(b_a, b_c) = \left| \mathbb{E}_{(x, y) \sim P_{b_a}}[\ell(f_\theta(x), y)] - \mathbb{E}_{(x, y) \sim P_{b_c}}[\ell(f_\theta(x), y)]\right|.
\end{equation}
By the Hoeffding's inequality, the following inequality holds with probability at least $1-\delta$, for all groups $b \in \mathcal{B}$,
\begin{equation}
\label{eq:Hoeffding}
    \left| \mathcal{L}_b - \hat{\mathcal{L}}_{{b}}\right| \leq C \cdot \sqrt{{2\log{(|\mathcal{B}|/\delta)}\over n_b}}.
    % \left| \mathbb{E}_{(x, y) \sim P_{b}}[\ell(f_\theta(x), y)] - \mathbb{E}_{(x, y) \sim \hat{P}_{b}}[\ell(f_\theta(x), y)]\right| \leq C \cdot \sqrt{{2\log{(|\mathcal{B}|/\delta)}\over n_b}}.
\end{equation}
Note that the loss $\ell(\cdot)$ is upper-bounded by some constant $C$ according to our assumption and is always non-negative.
Accordingly, the following holds with probability at least $1-\delta$,
\begin{align}
\label{eq:loss_diff_b0_b1}
    d(b_a, b_c) \leq &\left| \hat{\mathcal{L}}_{{b_a}} - \hat{\mathcal{L}}_{{b_c}} \right| + C \left( \sqrt{{2\log{(|\mathcal{B}|/\delta)}\over n_{b_a}}} + \sqrt{{2\log{(|\mathcal{B}|/\delta)}\over n_{b_c}}} \right).
    % d(b_a, b_c) \leq &\left| \mathbb{E}_{(x, y) \sim \hat{P}_{b_a}}[\ell(f_\theta(x), y)] - \mathbb{E}_{(x, y) \sim \hat{P}_{b_c}}[\ell(f_\theta(x), y)] \right| \\ &+ C \left( \sqrt{{2\log{(|\mathcal{B}|/\delta)}\over n_{b_a}}} + \sqrt{{2\log{(|\mathcal{B}|/\delta)}\over n_{b_c}}} \right). \nonumber
\end{align}
Since the equation $\max\{x, y\} = {x+y \over 2} + {|x - y| \over 2}$ holds for $x \in \mathbb{R}$ and $y \in \mathbb{R}$, the RHS of \Cref{eq:loss_diff_b0_b1} is at most:
\begin{align}
    \left| \hat{\mathcal{L}}_{{b_a}} - \hat{\mathcal{L}}_{{b_c}} \right| \leq 2 \cdot \max_{b\in\mathcal{B}} \hat{\mathcal{L}}_{{b}}. 
    % &\left| \mathbb{E}_{(x, y) \sim \hat{P}_{b_a}}[\ell(f_\theta(x), y)] - \mathbb{E}_{(x, y) \sim \hat{P}_{b_c}}[\ell(f_\theta(x), y)] \right| \\ &\leq 2 \cdot \max\{\mathbb{E}_{(x, y) \sim \hat{P}_{b_a}}[\ell(f_\theta(x), y)], \mathbb{E}_{(x, y) \sim \hat{P}_{b_c}}[\ell(f_\theta(x), y)]\}. \nonumber
\end{align}
Therefore, we have shown that the following result holds,
\begin{align}
    d(b_a, b_c) & \leq 2 \cdot \max_{b\in\mathcal{B}}\hat{\mathcal{L}}_{{b}} + C \left( \sqrt{{2\log{(|\mathcal{B}|/\delta)}\over n_{b_a}}} + \sqrt{{2\log{(|\mathcal{B}|/\delta)}\over n_{b_c}}} \right) \nonumber \\ & \leq 2 \cdot \max_{b\in\mathcal{B}}\hat{\mathcal{L}}_{{b}} + C \cdot \max_{b \in \mathcal{B}} \sqrt{{8\log{(|\mathcal{B}|/\delta)}\over n_{b}}}. \label{eq:theorem1}
    % d(b_a, b_c) & \leq 2 \cdot \max_{b\in\mathcal{B}}\mathbb{E}_{(x, y) \sim \hat{P}_{b}}[\ell(f_\theta(x), y)] + C \left( \sqrt{{2\log{(|\mathcal{B}|/\delta)}\over n_{b_a}}} + \sqrt{{2\log{(|\mathcal{B}|/\delta)}\over n_{b_c}}} \right) \nonumber \\ & \leq 2 \cdot \max_{b\in\mathcal{B}}\mathbb{E}_{(x, y) \sim \hat{P}_{b}}[\ell(f_\theta(x), y)] + C \cdot \max_{b \in \mathcal{B}} \sqrt{{8\log{(|\mathcal{B}|/\delta)}\over n_{b}}}. \label{eq:theorem1}
\end{align}
This completes the proof of \Cref{theorem1}.
\end{proof}

\renewcommand*{\proofname}{Proof of \Cref{theorem2}}
\begin{proof}
% \paragraph{Proof of Theorem 2.} 
% Considering that the training distribution is represented by a mixture of two group distributions $P_{b_a}$ and $P_{b_c}$ for two distinct groups $b_a \in \mathcal{B}$ and $b_c \in \mathcal{B}$, the average loss can be expressed as follows:
% \begin{equation}
%     \mathcal{L}_{\text{avg}} = k_{b_a} \cdot \mathcal{L}_{b_a} + k_{b_c} \cdot \mathcal{L}_{b_c},
%     % \mathcal{L}_{\text{avg}} = k_{b_a} \cdot \mathbb{E}_{(x, y) \sim P_{b_a}}[\ell(f_\theta(x), y)] + k_{b_c} \cdot \mathbb{E}_{(x, y) \sim P_{b_c}}[\ell(f_\theta(x), y)].
% \end{equation}
% where $k_{b_a}=P(b_a)$ and $k_{b_c}=P(b_c)$ denote the prior probability associated with groups. 
By the Hoeffding's inequality, the following inequality holds with at least $1-\delta$:
\begin{equation}
\label{eq:theorem2}
    \left|\mathcal{L}_{\text{avg}} - \hat{\mathcal{L}}_{\text{avg}}\right| \leq C \cdot \sqrt{2\log(1/\delta) \over n}.
\end{equation}
Accordingly, the expected average loss is bounded with probability $1-\delta$ as follows:
\begin{align}
    \hat{\mathcal{L}}_{\text{avg}} - C \cdot \sqrt{2\log(1/\delta) \over n} \leq \mathcal{L}_{\text{avg}} \leq \hat{\mathcal{L}}_{\text{avg}} + C \cdot\sqrt{2\log(1/\delta) \over n}.
\end{align}

Considering that the training distribution is represented by a mixture of two group distributions $P_{b_a}$ and $P_{b_c}$ for two distinct groups $b_a \in \mathcal{B}$ and $b_c \in \mathcal{B}$, $\hat{\mathcal{L}}_{\text{avg}}$ is equal to
\begin{align}
    k_{b_a} \cdot \hat{\mathcal{L}}_{{b_a}} + k_{b_c} \cdot \hat{\mathcal{L}}_{{b_c}} \leq \max_{b\in\mathcal{B}} \hat{\mathcal{L}}_{{b}},
    % k_{b_a} \cdot \mathbb{E}_{(x, y) \sim \hat{P}_{b_a}}[\ell(f_\theta(x), y)] + k_{b_c} \cdot \mathbb{E}_{(x, y) \sim \hat{P}_{b_c}}[\ell(f_\theta(x), y)] \leq \max_{b\in\mathcal{B}}\mathbb{E}_{(x, y) \sim \hat{P}_{b}}[\ell(f_\theta(x), y)].
\end{align}
where $k_{b_a}=P(b_a)$ and $k_{b_c}=P(b_c)$ denote the prior probability associated with groups.
Therefore, $\mathcal{L}_{\text{avg}}$ is upper-bounded with probability at least $1-\delta$:
\begin{equation}
    \mathcal{L}_{\text{avg}} \leq \max_{b\in\mathcal{B}}\hat{\mathcal{L}}_{{b}} + C \cdot \sqrt{2\log(1/\delta) \over n}.
    % \mathcal{L}_{\text{avg}} \leq \max_{b\in\mathcal{B}}\mathbb{E}_{(x, y) \sim \hat{P}_{b}}[\ell(f_\theta(x), y)] + C \cdot \sqrt{2\log(1/\delta) \over n}.
\end{equation}
This completes the proof of \Cref{theorem2}.
\end{proof}

\section{Experimental details}
\subsection{Benchmarks}
\label{apdx:dataset}
\begin{figure}[H]
\centering
\begin{minipage}{0.48\linewidth}
    \centering
    \includegraphics[width=66mm]{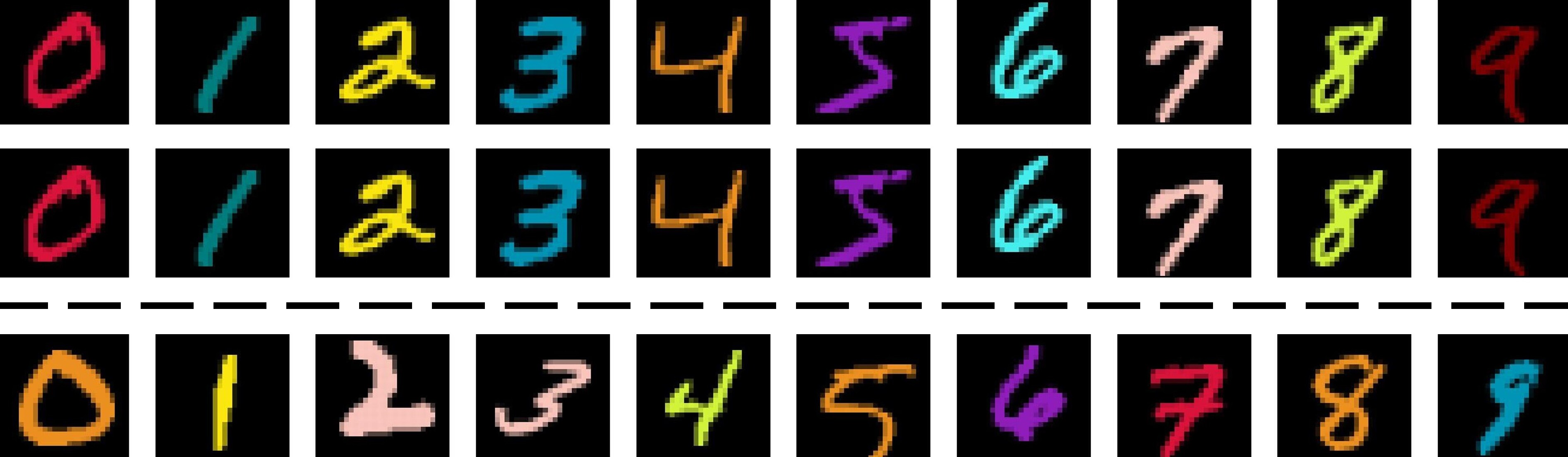}
    \caption{Colored MNIST.}\label{fig:cmnist}
\end{minipage}
% \hfill
\begin{minipage}{0.48\linewidth}
    \centering
    \includegraphics[width=66mm]{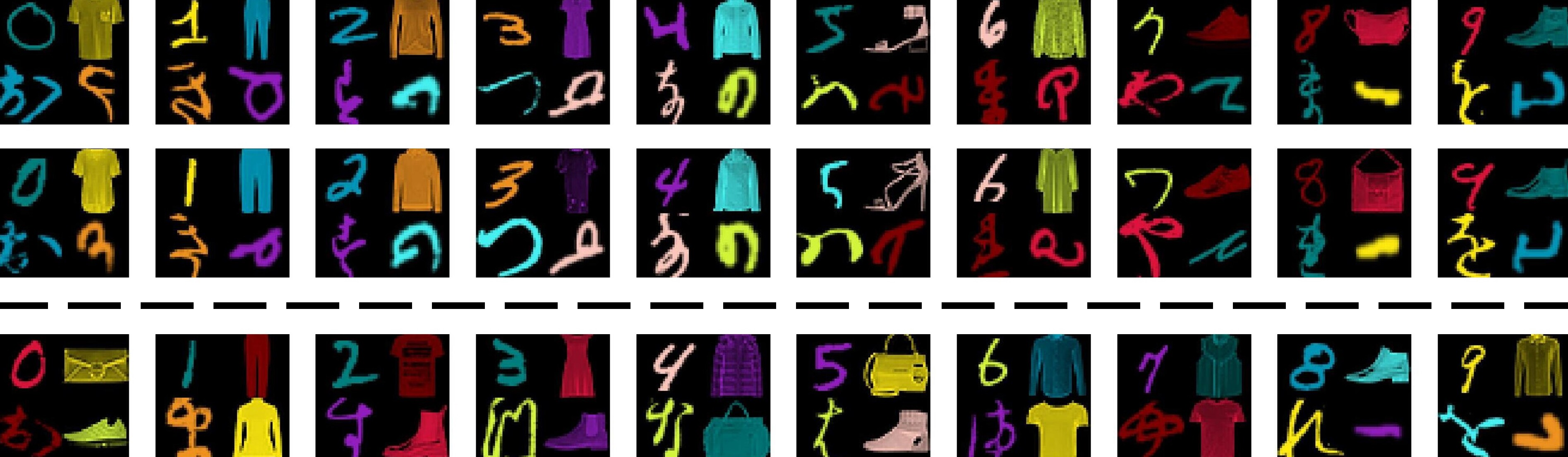}
    \caption{Multi-bias MNIST.}\label{fig:mbmnist}
\end{minipage}
\end{figure}

\begin{figure}[H]
\centering
\begin{minipage}{0.48\linewidth}
    \centering
    \includegraphics[width=65mm]{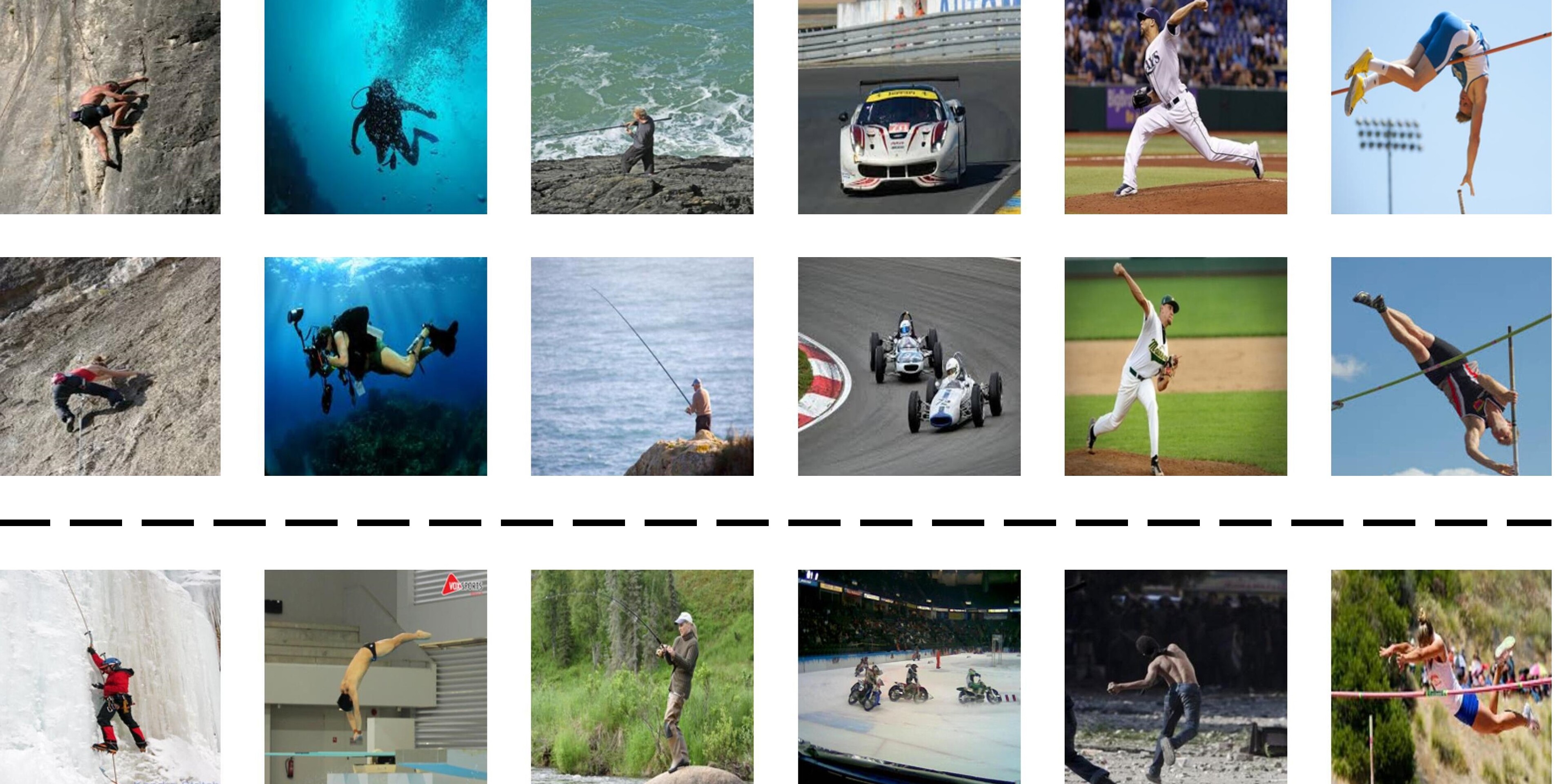}
    \caption{Biased action recognition.}\label{fig:bar}
\end{minipage}
\begin{minipage}{0.245\linewidth}
    \centering
    \includegraphics[width=31.6mm]{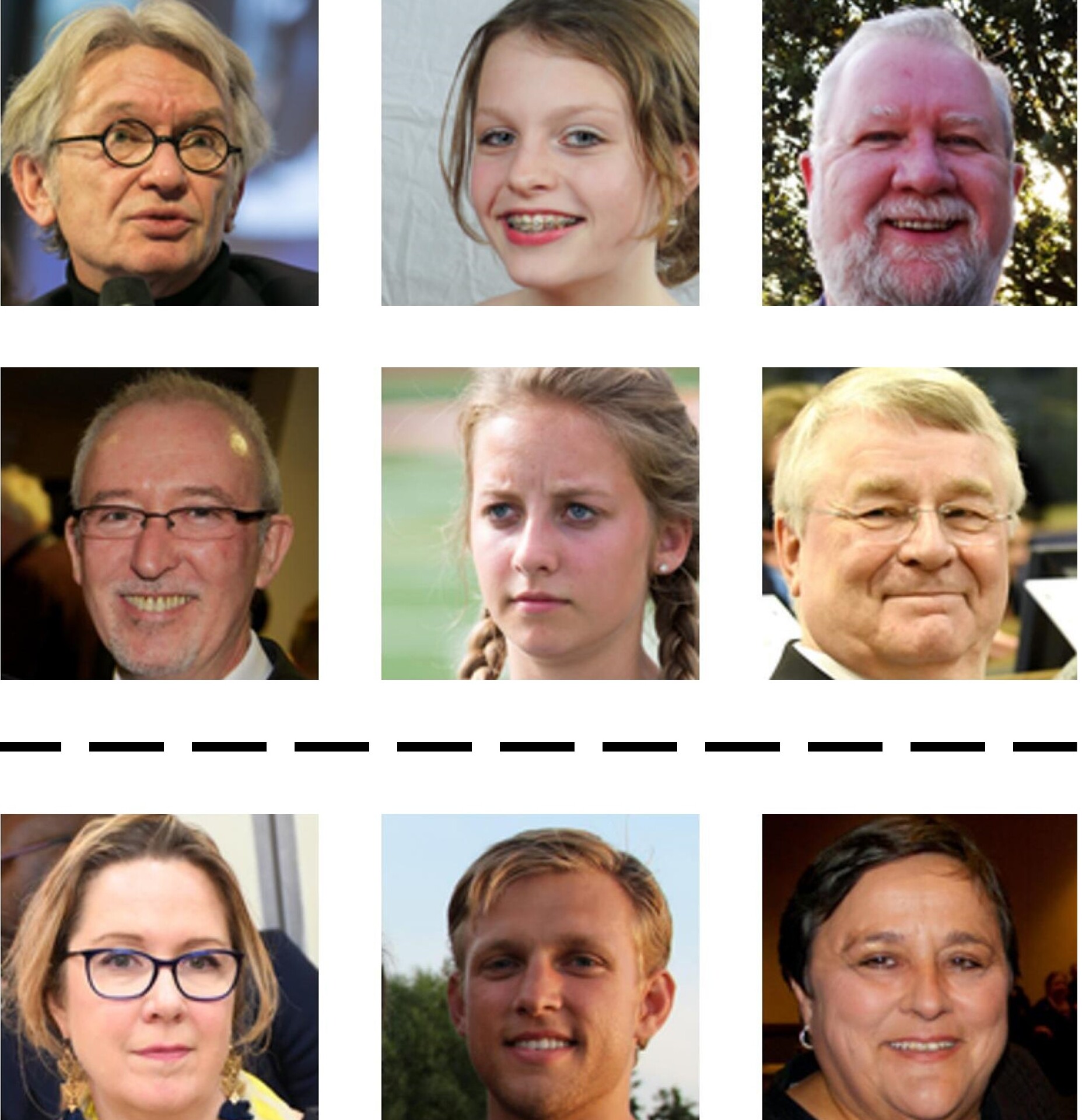}
    \caption{Biased FFHQ.}\label{fig:bffhq}
\end{minipage}
\begin{minipage}{0.245\linewidth}
    \centering
    \includegraphics[width=31.6mm]{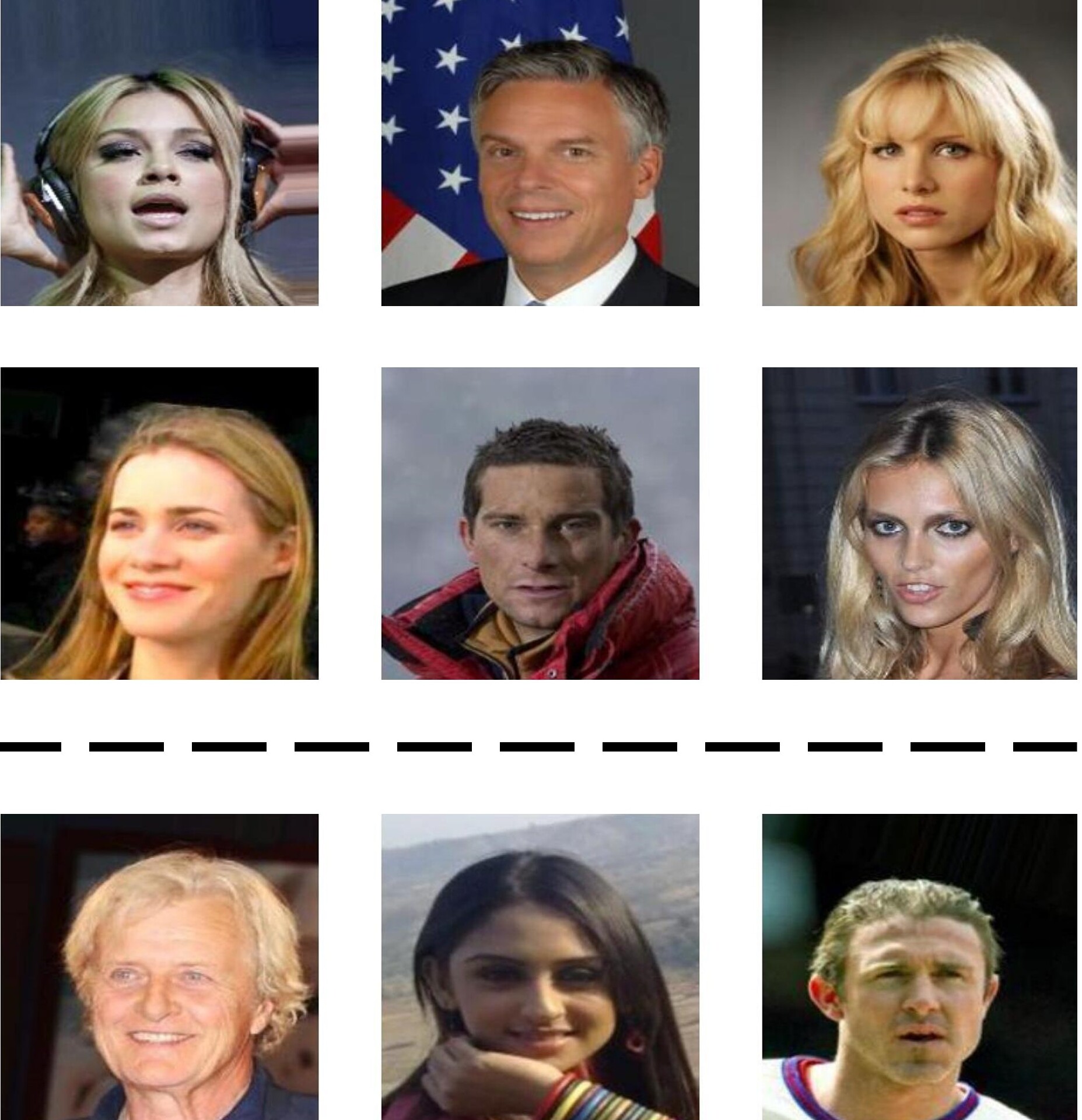}
    \caption{CelebA.}\label{fig:celeba}
\end{minipage}
\end{figure}

We provide a detailed description of the datasets utilized in \Cref{sc:exp}. 
In \Cref{fig:cmnist,fig:mbmnist,fig:bar,fig:bffhq,fig:celeba}, each column corresponds to a distinct class. The images positioned above the dashed line represent bias-aligned samples, whereas those below represent bias-conflicting samples.
\paragraph{Colored MNIST.}
Colored MNIST (C-MNIST) is a synthetic dataset designed for digit classification that introduces a controlled spurious correlation between digit shape and color. This dataset consists of ten digits from the MNIST dataset, each associated with a specific RGB color. Following the setting in \citet{PGD}, ten RGB colors are uniformly selected and applied to the grayscale MNIST images. The image generation involves assigning colors based on the predetermined bias-conflicting ratio, $\rho$. Specifically, each image is colored to create a bias-conflicting sample with probability $\rho$, and a bias-aligned sample otherwise. For bias-aligned samples of a given class $y$, the digit is colored using $c \sim \mathcal{N}(C_y, \sigma I_{3 \times 3})$, where $C_y \in \mathbb{R}^3$ is the predefined color vector for class $y$. Conversely, bias-conflicting samples receive a color from any class other than $y$, i.e. $C_{U_y} \in \{C_i\}_{i \in [10]\setminus y}$, and colored using $c \sim \mathcal{N}(C_{U_y}, \sigma I_{3 \times 3})$. The experiments utilize bias-conflicting ratios of 0.5\%, 1\%, and 5\%, with $\sigma$ set to 0.0001. Additionally, we use a 10\% of training data as validation data, and an unbiased test set with a bias-conflicting ratio of 90\% is employed for performance evaluation.

\paragraph{Multi-bias MNIST.}
Multi-bias MNIST (MB-MNIST) is a synthetic dataset derived from \citet{PGD} to incorporate more complex patterns compared to C-MNIST and biased-MNIST \cite{biased_mnist}. MB-MNIST comprises eight attributes: digit \cite{mnist}, alphabet \cite{emnist}, fashion object \cite{fmnist}, Japanese character \cite{japanese}, digit color, alphabet color, fashion object color, and Japanese character color. The digit shape is the target attribute, while the other seven attributes serve as bias attributes. In MB-MNIST, bias is introduced by aligning the digit and its color with probability $1-\rho$. Similarly, the other six attributes are also independently aligned with the digit, each with probability $1-\rho$. In our setting, the ratios of bias-conflicting samples are $10\%$, $20\%$, and $30\%$ as in \citet{PGD}. As CMNIST, we use 10\% of the training data as validation data. Test samples are generated with a bias-conflicting ratio of 90\% for all bias attributes to create an unbiased test set.

\paragraph{Biased action recognition.} 
The biased action recognition (BAR) dataset \cite{LfF}, designed for action classification, is a real-world dataset that encompasses six action classes: climbing, diving, fishing, racing, throwing, and vaulting. Each class is spuriously correlated with a specific place. The training set of BAR consists exclusively of bias-aligned samples, while the test set consists solely of bias-conflicting samples. For example, all the training samples for the climbing class are associated with rock walls, but the test samples for climbing are associated with different settings, such as ice cliffs. We utilize the data splits defined in \citet{PGD}, which allocate 10\% of the training set to the validation set.

\paragraph{Biased FFHQ.}
Biased FFHQ (BFFHQ) was constructed from the real-world facial dataset FFHQ \cite{ffhq}, which has age (young or old) as a label and gender (male or female) as a bias feature. This benchmark was conducted in previous works \cite{DFA,PGD,biaswap}. Individuals aged between 10 and 29 are labeled as ``young’’, while those aged between 40 and 59 are labeled as ``old''. In this dataset, the majority of females are young, while males are predominantly old. The training set contains only 0.5\% bias-conflicting samples. We report accuracies on both an unbiased test set with a bias-conflicting ratio of 50\% \cite{PGD} and a bias-conflicting test set \cite{DFA}.

\paragraph{CelebA.}
CelebA \cite{CelebA} is a real-world dataset for face classification. Following \citet{groupDRO}, the goal is to classify celebrities’ hair color as either blond or non-blond, which exhibits a spurious correlation with gender (male or female). Notably, only a small fraction of blond-haired celebrities are male, which leads to poor performance of ERM-trained models on this group. We employ the training, validation, and test splits as specified in \citet{groupDRO} and report both the average accuracy and the worst-group accuracy on the test set. In this context, the group is defined as the combination of the class label and the bias label as in prior studies \cite{CNC,LCloss,groupDRO}.

\paragraph{CivilComments-WILDS.}
CivilComments-WILDS \cite{civilcomments_borkan,wilds} is a dataset for text classification problems. The goal is to classify whether an online comment is toxic or non-toxic, which exhibits a spurious correlation with the mention of certain demographic identities, including male, female, white, black, LGBTQ, Christian, Muslim, and other religions. We use the same data splits as described in \citet{wilds} and report both the average accuracy and the worst-group accuracy on the test dataset. Here, groups are defined as combinations of class labels and bias labels, as described in previous works \cite{JTT,wilds,CNC}.

% \subsection{Model architectures and hyperparameters}
\subsection{Implementation details}
\label{apdx:impdetail}
We provide descriptions of the implementation details. For CelebA and CivilComments-WILDS, we follow the experimental settings outlined in \citet{CNC} and \citet{JTT}, respectively. For the other datasets—CMNIST, MB-MNIST, BAR, and BFFHQ—we follow the experimental setups presented in \citet{PGD}. The hyperparameter for GCE, denoted as $q$, is set to 0.7 as in \citet{GCE}. All classification models are trained using an NVIDIA RTX A6000.

\subsubsection{Model architectures and hyperparameters}
The model architectures and hyperparameters for each dataset are described:
\paragraph{C-MNIST.}
As in \citet{PGD}, we utilize a simple CNN architecture (SimConv-1). Please see Appendix B of \citet{PGD} for detailed network implementation. We train the model for 100 epochs with SGD optimizer, a batch size of 128, a learning rate of 0.02, weight decay of 0.001, momentum of 0.9, and learning rate decay of 0.1 at every 40 steps. For C-MNIST with bias-conflicting ratios of 0.5\% and 1\%, we use a temperature of 1; for a ratio of 5\%, we use a temperature of 1.1.

\paragraph{MB-MNIST.} 
We use a different type of simple CNN model (SimConv-2), following \citet{PGD}. Please refer to Appendix B in \citet{PGD} for network implementation. We train for 100 epochs with SGD optimizer, a batch size of 32, a learning rate of 0.01, and weight decay of 1e-4, momentum of 0.9. 
For the MB-MNIST dataset, we use temperatures of 0.9, 1.1, and 1.3 for the ratios of 10\%, 20\%, and 30\%, respectively.

\paragraph{BAR.}
We use a ResNet18 pretrained on ImageNet as in \citet{biaswap}. We train for 100 epochs with SGD optimizer, a batch size of 16, a learning rate of 5e-4, weight decay of 1e-5, momentum of 0.9, learning rate decay of 0.1 at every 20 steps, and a temperature of 0.6.

\paragraph{BFFHQ.}
We utilize an ImageNet-pretrained ResNet18 as in \citet{DFA}. We train for 160 epochs with Adam optimizer, a batch size of 64, a learning rate of 1e-4, weight decay of 0, learning rate decay of 0.1 at every 32 steps, and a temperature of 0.9.

\paragraph{CelebA.}
We utilize a ImageNet-pretrained ResNet50 and hyperparameters from \citet{CNC}. We train for 50 epochs with SGD optimizer, a batch size of 128, a learning rate of 1e-4, weight decay of 0.1, and a temperature of 1.

\paragraph{CivilComments-WILDS.}
We utilize a pretrained BERT with the number of tokens capped at 300 following previous works \cite{CNC,JTT,PGD,wilds}. We train the biased model using the SGD optimizer without gradient clipping. In contrast, for training the debiased model, we employ the AdamW optimizer with gradient clipping and a linearly-decaying learning rate. Both models are trained for up to 5 epochs with a batch size of 16, a learning rate of 1e-5, weight decay of 0.01, and a temperature of 2.

\subsubsection{Image preprocessing}
We employ the same image preprocessing scheme as presented in \citet{PGD}. For CMNIST and MB-MNIST, we use fixed sizes of (28×28) and (56×56), respectively. For the remaining datasets, we use a fixed size of (224×224). Additionally, we normalize images from BAR and BFFHQ with a mean of (0.4914, 0.4822, 0.4465) and a standard deviation of (0.2023, 0.1994, 0.2010). For vision datasets except for CelebA, we apply random resize crop, random color jitter, and random rotation to increase the diversity of bias-conflicting samples. To ensure a fair comparison, the same data augmentation techniques are applied to both the baselines and the proposed DPR.

% \subsection{Implementation}

\section{Additional ablation study}
\subsection{Ablation study on $q$ of GCE}
\begin{table}[h!]
% \begin{wraptable}{r}{0.5\textwidth} % "r" for right side, "l" for left side
\centering
\vspace{-4.0mm}
\caption{Ablation study on $q$ of GCE.}
\label{table:ablt-q}
\resizebox{0.62\columnwidth}{!}{%
\begin{tabular}{ccccc}
\toprule[1pt]
\multicolumn{1}{l}{} & \multicolumn{2}{c}{C-MNIST (0.5\%)} & \multicolumn{2}{c}{MB-MNIST (30\%)} \\ \cmidrule(lr){2-3} \cmidrule(lr){4-5} 
\multicolumn{1}{l}{} & Biased model    & Debiased model    & Biased model    & Debiased model    \\ \midrule
$q = 0.3$            & 26.27 (2.11)    & 95.53 (0.74)      & 79.73 (0.91)    & 95.32 (0.71)      \\
$q = 0.5$            & 21.30 (3.10)    & 96.83 (0.32)      & 83.16 (0.56)    & 94.84 (0.15)      \\
$q = 0.7$            & 18.55 (2.91)    & 97.52 (0.33)      & 85.47 (1.26)    & 94.04 (0.26)      \\
$q = 0.9$            & 16.19 (0.46)    & 97.66 (0.06)      & 85.26 (0.70)    & 93.77 (0.19)      \\ \bottomrule[1pt]
\end{tabular} }
% \end{wraptable}
\end{table}
We conduct an ablation study on the C-MNIST and MB-MNIST datasets, which have bias-conflicting ratios of 0.5\% and 30\%, respectively, to assess the impact of GCE parameter $q$. As depicted in \Cref{table:ablt-q}, varying $q$ demonstrates distinct effects on the performance of biased and debiased models. For C-MNIST, increasing $q$ enhances the debiased model's performance while degrading that of the biased model. In contrast, for MB-MNIST, decreasing $q$ enhances the performance of the debiased model but worsens the performance of the biased model. These results suggest that the optimal setting of $q$ varies between datasets. Additionally, a consistent observation from both datasets is that as the performance of the biased model decreases, the performance of the debiased model increases. This trend implies that a lower performance of the biased model, indicative of a stronger focus on spurious correlations, leads to more accurate group proxies, which in turn boosts the debiased model’s performance.

\subsection{Ablation study on calibration hyperparameter $\tau$}
\begin{table}[h!]
% \begin{wraptable}{r}{0.5\textwidth} % "r" for right side, "l" for left side
\centering
\vspace{-4.0mm}
\caption{Ablation study on calibration hyperparameter $\tau$.}
\label{table:ablt-tau}
\resizebox{0.55\columnwidth}{!}{%
\begin{tabular}{ccccc}
\toprule[1pt]
\multicolumn{1}{l}{} & \multicolumn{2}{c}{C-MNIST}                   & \multicolumn{2}{c}{MB-MNIST}                  \\ \cmidrule(lr){2-3} \cmidrule(lr){4-5} 
Ratio (\%)           & 0.5                   & 5                     & 20                    & 30                    \\ \midrule
$\tau = 0.9$         & 97.48 (0.41)          & 97.96 (0.28)          & 88.10 (1.78)          & 91.53 (0.87)          \\
$\tau = 1.0$         & \textbf{97.52 (0.33)} & 98.38 (0.16)          & 88.68 (1.69)          & 92.85 (0.65)          \\
$\tau = 1.1$         & 96.48 (0.16)          & \textbf{98.62 (0.12)} & \textbf{89.11 (1.65)} & 92.99 (0.66)          \\
$\tau = 1.2$         & 95.84 (0.46)          & 98.40 (0.18)          & 87.80 (2.49)          & 93.49 (0.47)          \\
$\tau = 1.3$         & 95.07 (0.36)          & 98.42 (0.02)          & 87.73 (1.53)          & \textbf{94.04 (0.26)} \\ \bottomrule[1pt]
\end{tabular} }
% \end{wraptable}
\end{table}
Since DPR utilizes the calibration hyperparameter $\tau$ for capturing the spurious correlation structure well, we conduct an ablation study on C-MNIST and MB-MNIST across various bias-conflicting ratios to assess the effect of $\tau$. \Cref{table:ablt-tau} illustrates how performance varies with different settings of $\tau$. For C-MNIST, the best performances are achieved at $\tau=1$ and $\tau=1.1$ for bias-conflicting ratios of 0.5\% and 5\%, respectively. For MB-MNIST, the best results are obtained with $\tau=1.1$ and $\tau=1.3$ for ratios of 20\% and 30\%, respectively. The optimal $\tau$ can vary not only between different datasets but also according to the extent of the prevalence of spurious correlations within the datasets. These results suggest that adjusting $\tau$ is effective in capturing the diverse spurious correlation structures.

\subsection{Comparison of resampling and reweighting}
\begin{table}[h!]
% \begin{wraptable}{r}{0.40\textwidth} % "r" for right side, "l" for left side
\centering
\vspace{-4.0mm}
\caption{Comparison of resampling and reweighting.}
\label{table:weight-sample}
\resizebox{0.47\columnwidth}{!}{%
\begin{tabular}{cccc}
\toprule[1pt]
\multicolumn{1}{l}{} & \multicolumn{3}{c}{C-MNIST}                \\ \cmidrule(lr){2-4}
Ratio (\%)           & 0.5          & 1            & 5            \\ \midrule
Resampling           & 97.52 (0.33) & 98.40 (0.03) & 98.62 (0.12) \\
Reweighting          & 95.04 (0.35) & 96.21 (0.45) & 97.84 (0.50) \\ \bottomrule[1pt]
\end{tabular} }
% \end{wraptable}
\end{table}
While reweighting could effectively solve the weighted loss minimization problem presented in \Cref{eq:reweighted}, DPR adopts resampling. To justify this choice, we conduct experiments to compare resampling with reweighting. \Cref{table:weight-sample} shows the performance of each method on the C-MNIST dataset across various bias-conflicting ratios. The results reveal that DPR using the resampling strategy consistently outperforms DPR using the reweighting approach. This superiority of resampling over reweighting has been explored and explained by \citet{an2021}.

\end{document}